\documentclass[10pt,twocolumn,letterpaper]{article}

\usepackage{cvpr}              
\usepackage{multirow}
\usepackage{makecell}
\usepackage{xcolor}
\usepackage{graphicx}
\definecolor{orange}{HTML}{DB5C4A}
\definecolor{red}{HTML}{C00000}
\definecolor{blue}{HTML}{215F9A}
\definecolor{failure}{HTML}{B5525D}
\definecolor{coolwarm}{HTML}{D95747}
\usepackage[table]{xcolor} 
\newcommand{\tablestyle}[2]{\setlength{\tabcolsep}{#1}\renewcommand{\arraystretch}{#2}\centering\footnotesize}
\definecolor{cvprblue}{rgb}{0.21,0.49,0.74}
\usepackage[pagebackref,breaklinks,colorlinks,allcolors=cvprblue]{hyperref}

\def\paperID{*****}
\def\confName{CVPR}
\def\confYear{2026}

\title{
TALO: Pushing 3D Vision Foundation Models Towards \\Globally Consistent Online Reconstruction
}

\author{
Fengyi Zhang\textsuperscript{1}\;
Tianjun Zhang\textsuperscript{2}\;
Kasra Khosoussi\textsuperscript{1}\;
Zheng Zhang\textsuperscript{3}\;
Zi Huang\textsuperscript{1}\;
Yadan Luo\textsuperscript{1}\thanks{Corresponding Author.}\\
{\textsuperscript{1}UQMM Lab, The University of Queensland} \;
{\textsuperscript{2}Shanghai Jiao Tong University} \\
{\textsuperscript{3}Harbin Institute of Technology}\\
}

\begin{document}

\maketitle
\begin{abstract}
3D vision foundation models have shown strong generalization in reconstructing key 3D attributes from uncalibrated images through a single feed-forward pass. However, when deployed in online settings such as driving scenarios, predictions are made over temporal windows, making it non-trivial to maintain consistency across time. Recent strategies align consecutive predictions by solving global transformation, yet our analysis reveals their fundamental limitations in assumption validity, local alignment scope, and robustness under noisy geometry. In this work, we propose a higher-DOF and long-term alignment framework based on Thin Plate Spline, leveraging globally propagated control points to correct spatially varying inconsistencies. In addition, we adopt a point-agnostic submap registration design that is inherently robust to noisy geometry predictions. The proposed framework is fully plug-and-play, compatible with diverse 3D foundation models and camera configurations (e.g., monocular or surround-view). Extensive experiments demonstrate that our method consistently yields more coherent geometry and lower trajectory errors across multiple datasets, backbone models, and camera setups, highlighting its robustness and generality. 
Code is available at \url{https://github.com/Xian-Bei/TALO}.  
\end{abstract}

\vspace{-1ex}
\section{Introduction}
\label{sec:intro}\vspace{-1ex}

\begin{figure*}[t]
  \centering
  \includegraphics[width=\linewidth]{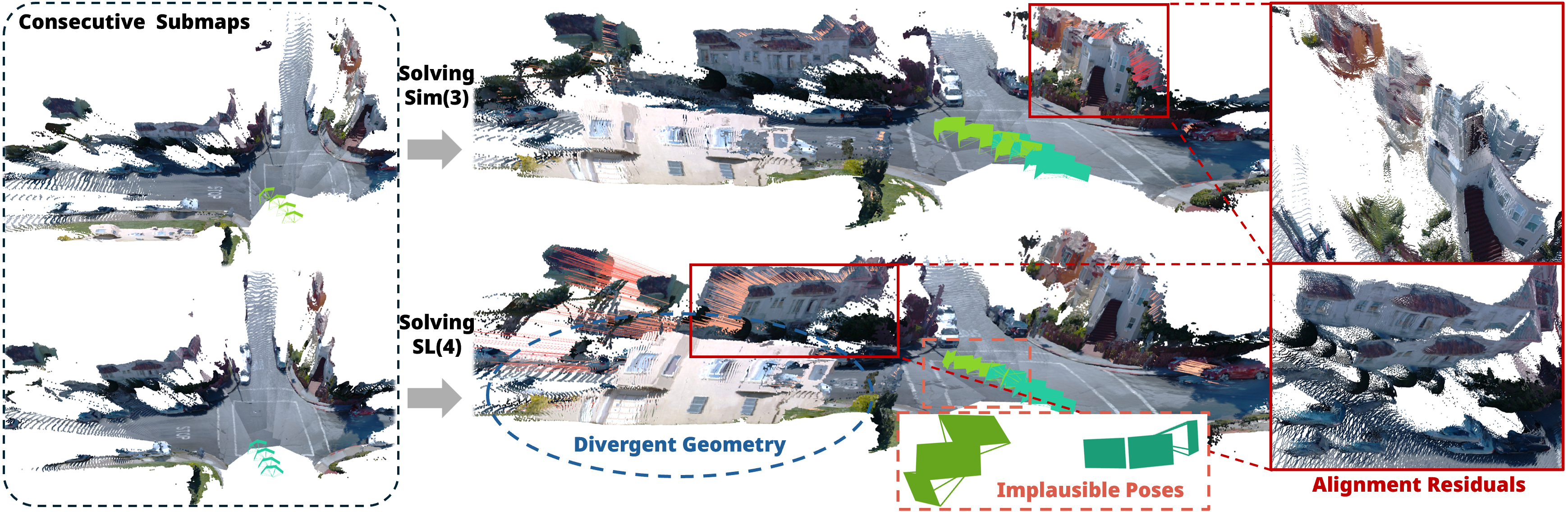}\vspace{-0.5ex}
   \caption{
\textbf{Degeneration of $\mathrm{Sim}(3)$ alignment used in VGGT-Long \cite{vggtlong} and $\mathrm{SL}(4)$ in VGGT-SLAM \cite{vggtslam}.}
Two consecutive submaps (left) independently predicted by a foundation model exhibit spatially varying, nonlinear geometric inconsistencies.
Since neither $\mathrm{Sim}(3)$ nor $\mathrm{SL}(4)$ can theoretically reconcile such non-global distortions with a single global transformation, enforcing a global warp inevitably overfits one region at the expense of another, leaving noticeable alignment residuals (visualized by \textcolor{coolwarm}{colored connecting lines} and \textcolor{red}{\textbf{zoomed-in}} on the right, where severe wall ghosting occurs).
Moreover, the under-constrained $\mathrm{SL}(4)$ is highly sensitive to noise and often yields divergent geometry (e.g., severely tilted buildings, \textcolor{blue}{\textbf{blue circle}}) and physically implausible camera poses.
As shown in the \textcolor{orange}{\textbf{orange box}}, the three cameras predicted by the foundation model on the right maintain consistent forward-facing orientations, whereas after $\mathrm{SL}(4)$ alignment, their pitch angles diverge drastically,
resulting in an impossible trajectory in driving scenarios.
}
    \vspace{-1.5ex}
   \label{fig:sim3sl4}
\end{figure*}




Recent progress in 3D vision foundation models (3DVFMs) such as VGGT \cite{vggt}, $\pi^3$ \cite{pi3} and MapAnything \cite{mapanything} has redefined the paradigm of 3D reconstruction. 
These models estimate camera intrinsics, poses and dense geometry directly from uncalibrated images through a single feed-forward inference, showing strong generalization across diverse scenarios. 
Nevertheless, most 3DVFMs are designed for \emph{offline reconstruction}, where the full sequence is processed at once. When applied to \emph{online settings} such as autonomous driving \cite{end2endad,ttocc,3ddet,3ddet2} and robotic manipulation \cite{robotic,robotic1,agent,agent2}, predictions are made per local window (\textit{i.e.,} submaps), making it non-trivial to maintain consistency across submaps since each is inferred independently.

A common remedy is to align consecutive submaps by solving a \emph{global} transformation between their overlapping regions. 
VGGT-Long \cite{vggtlong} adopts a stable $\mathrm{Sim}(3)$ alignment between submaps, which constrains the solution to be 7-DOF to account for rotation, translation, and a global scale. 
VGGT-SLAM \cite{vggtslam} further observes that 3DVFMs often produce \emph{inconsistent intrinsics} across submaps, for which a single $\mathrm{Sim}(3)$ warp is \emph{insufficient} to successfully align consecutive point clouds. 
To address this, VGGT-SLAM replaces $\mathrm{Sim}(3)$ with a 15-DOF $\mathrm{SL}(4)$ warp, improving alignment in controlled indoor datasets. 
However, our empirical study indicates that $\mathrm{SL}(4)$ remains highly unstable and fragile in outdoor multi-camera settings, diverging in over 60\% of tested scenes across three foundation models \cite{vggt,pi3,mapanything}.

Our analysis identifies three fundamental limitations of these approaches, as illustrated in Fig.~\ref{fig:sim3sl4}.  
The first is their implicit assumption of \emph{globally uniform error fields} that can be corrected by a single linear transformation (see supplementary materials for a formal derivation).
Such an assumption is easily violated in realistic outdoor scenarios, especially with multiple cameras and small submap sizes, where geometric distortions \emph{vary spatially}.  
For example, consecutive submaps may exhibit opposite depth-scale and field-of-view biases across different cameras.
In such cases, enforcing a \emph{global} warp, whether $\mathrm{Sim}(3)$ or $\mathrm{SL}(4)$, inevitably overfits one region at the expense of another, leaving noticeable residuals and compromising trajectory accuracy to compensate for \emph{local} inconsistencies.

Moreover, current pipelines only perform \emph{pairwise alignment} between neighboring submaps, which can only guarantee \emph{short-term optimality}.  
Before loop closure is triggered, no information from distant submaps is utilized, limiting their ability to achieve global consistency and making them prone to trajectory drift, particularly in long-range sequences.  
Finally, the under-constrained nature of $\mathrm{SL}(4)$ makes it highly sensitive to geometry noise from 3DVFMs' predictions, often leading to degenerate solutions of implausible poses and divergent scene structures.

These observations motivate our solution \textbf{TALO}, a higher-DOF \underline{t}hin-plate \underline{a}lignment framework that is capable of leveraging \underline{l}ong-term information to correct spatially varying geometric inconsistencies across submaps in an \underline{o}nline manner.   In addition, we introduce a point-agnostic submap registration strategy that remains inherently robust to the noisy geometry predicted by 3DVFMs.

Specifically, TALO employs a sparse set of 3D control points that are uniformly distributed to ensure broad spatial coverage. To establish global connections among submaps, these control points are temporally propagated both forward and backward along the sequence, accumulating rich multi-view observations.
Each control point aggregates its multi-camera observations through robust fusion to suppress the influence of dynamic objects and outliers, yielding a globally consistent estimate of its canonical 3D position.  
Based on these correspondences, Thin Plate Spline (TPS) \cite{tps1,tps2,tps3,tps4} deformation fields are fitted to warp submaps into a shared canonical configuration, enforcing global geometric consistency.
TPS offers flexible, spatially varying correction while regularizing toward local rigidity to preserve structural coherence within each submap.  
Moreover, we register consecutive submaps by averaging the relative transformations of overlapping frames rather than optimizing over noisy point clouds.
This principled, point-agnostic formulation has been extensively validated to produce trajectory estimates that remain robust against geometry noise from 3DVFMs. 
Our main contributions are as follows:
\begin{itemize}
\item
We provide a systematic analysis of existing 3DVFM alignment strategies, revealing their fundamental limitations in assumption validity, local alignment scope, and robustness under noisy geometry.


\item
We propose TALO, a Thin Plate Spline-based alignment framework that leverages global information to correct spatially varying inconsistencies, together with a point-agnostic registration design robust to noisy predictions.

\item
We deliver a comprehensive, \emph{plug-and-play} system that seamlessly supports any foundation models (\textit{e.g.,} VGGT \cite{vggt}, $\pi^3$ \cite{pi3}, or MapAnything \cite{mapanything}) and arbitrary camera settings (\textit{e.g.,} monocular or surround-view).

\item
Extensive experiments demonstrate that TALO consistently produces more coherent geometry and reduced trajectory errors across all foundation models, datasets, and camera setups, highlighting its robustness and generality.
\end{itemize}


\section{Related Work}
\subsection{Multi-Stage 3D Reconstruction}
Conventional 3D reconstruction pipelines typically begin with Structure-from-Motion (SfM), which recovers only sparse geometry and camera parameters.  
Classical SfM systems \cite{Openmvg,colmap1,sfm1,sfm2,sfm3} generally follow a sequence of stages, including feature extraction, cross-view matching, triangulation, and bundle adjustment.
Dense geometry is then reconstructed using Multi-View Stereo (MVS) \cite{mvs1,mvs2,mvs3,mvs4}, or more recent neural approaches such as Neural Radiance Fields \cite{nerf,nerfplus,nerfstudio,ngp} and 3D Gaussian Splatting \cite{3dgs,2dgs,gforest,MipSplatting,GSLAM}, which typically rely on the sparse structure and camera poses estimated by SfM.  
To improve robustness, a number of learning-based approaches have been proposed to replace either individual modules \cite{loftr,superglue,deepsfm,banet,idacs} or the entire pipeline \cite{vggsfm,deeptwoview,colmapfree,colmapfree2}, making the process differentiable and able to benefit from learned data priors.  
Nevertheless, they remain fundamentally multi-stage and often rely on iterative optimization.

\subsection{Feed-Forward 3D Reconstruction}
A paradigm shift toward end-to-end, feed-forward 3D reconstruction was introduced by DUSt3R \cite{DUSt3R}, which predicts dense pointmaps directly from pairs of RGB images, thereby bypassing the traditional multi-stage geometry pipeline.  
Subsequent works \cite{mast3r,spann3r,fast3r,must3r,mvdust3r,monst3r} generalize this idea along various directions, including scalability to larger sets, handling dynamic scenes, and improving accuracy.  
Moving beyond predicting dense geometry alone, VGGT \cite{vggt} proposes a unified transformer architecture capable of predicting all core 3D attributes in a single forward pass.  
$\pi^{3}$ \cite{pi3} further introduces a permutation-equivariant design that removes reference-view bias, while MapAnything \cite{mapanything} extends toward a universal model that supports heterogeneous input–output modality combinations and metric-scale geometry.  
Together, these feed-forward models unify the prediction of all core 3D attributes within a single network, redefining 3D reconstruction as an end-to-end, single-shot inference problem.
We collectively refer to them as \emph{3D Vision Foundation Models (3DVFMs)}.

\subsection{Online 3D Reconstruction}
3DVFMs are generally designed to process two images or an entire sequence at once, requiring additional mechanisms for online reconstruction.  
Stream3R \cite{stream3r}, CUT3R \cite{cut3r}, and SLAM3R \cite{slam3r} are built upon DUSt3R \cite{DUSt3R} / MASt3R \cite{mast3r}:  
Stream3R introduces a causal decoder-only Transformer with KV-cached attention for long-sequence processing;  
CUT3R maintains a persistent state that continuously updates without restarting inference;  
and SLAM3R performs local multi-view reconstruction and progressive global registration without explicit pose estimation.  
StreamVGGT \cite{streamVGGT}, built upon VGGT \cite{vggt}, employs causal attention and token memory for sequential video processing.  
Beyond architectural extensions, alignment-based methods explicitly address cross-submap consistency and have demonstrated improved stability in practice.
VGGT-SLAM \cite{vggtslam} formulates submap alignment on the $\mathrm{SL}(4)$ manifold,
while VGGT-Long \cite{vggtlong} adopts low-DOF $\mathrm{Sim}(3)$ to enable robust long-range reconstruction; both follow a global alignment paradigm.
Our approach builds upon this explicit alignment paradigm, introducing a more flexible and robust design.



\section{Method}




\subsection{TALO Overview}

\begin{figure*}[t]
  \centering
  \includegraphics[width=\linewidth]{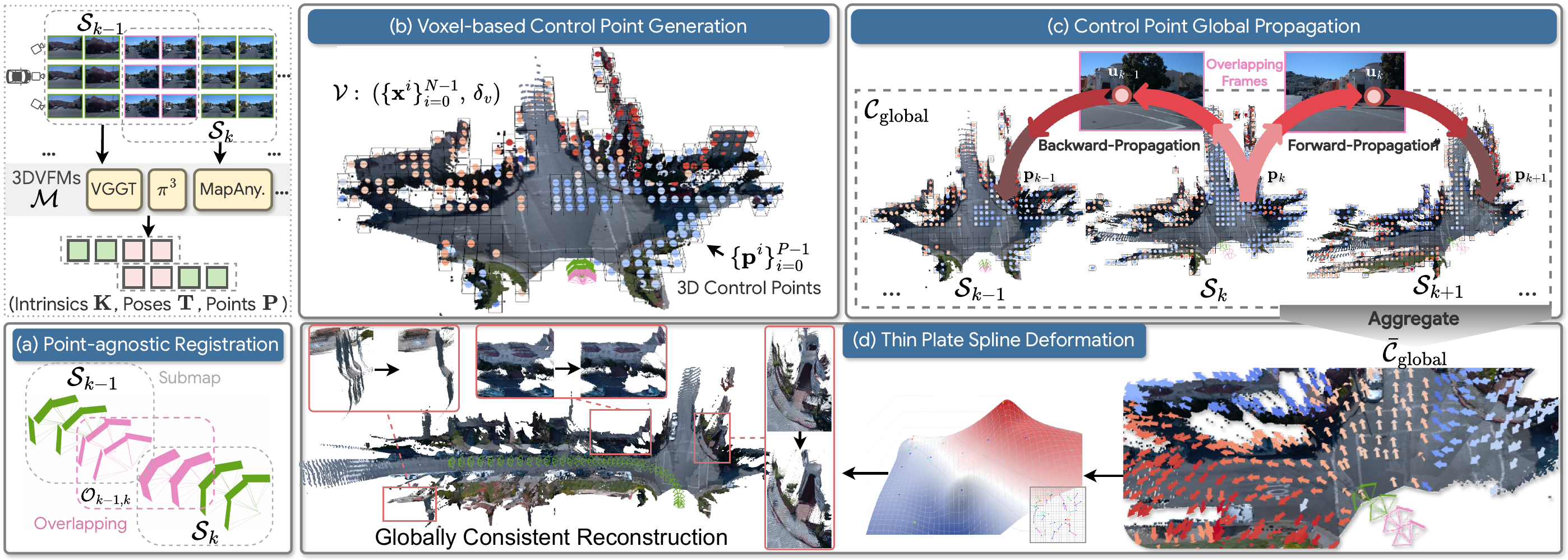}\vspace{-1ex}
   \caption{\textbf{Workflow of TALO}.
TALO processes multi-camera continuous sequences by first dividing them into submaps with overlapping frames. It then performs point-agnostic submap registration by averaging the relative transformations of the overlapping frames, followed by generating control points within the overlapping regions. These control points are globally propagated across submaps, and all their observations are finally aggregated to construct a TPS deformation field that warps every submap into a globally consistent canonical space.}\vspace{-2ex}
   \label{fig:overview}
\end{figure*}

To contextualize our framework, we first clarify the notation hierarchy used throughout this paper.
A \textbf{frame} refers to a synchronized set of \textbf{images} captured by different cameras at a single time step $t$.
A \textbf{submap} denotes a short temporal segment composed of several consecutive \textbf{frames}.
Since our system supports arbitrary $C$-camera configurations, all formulations are presented in a generic form applicable to monocular, stereo, and surround-view setups, and the index range $c \in \{0,\ldots C-1\}$ is omit for simplicity. 
We define the first camera ($c = 0$) of each frame as the reference camera.

Given a continuous, synchronized multi-camera video stream 
$\{ \mathbf{I}^{t,c}\}_{t=0}^{T-1}$,
we segment it into $K$ submaps $\{\mathcal{S}_k\}$, each containing $O$ overlapping frames with the previous submap and $L$ newly observed frames ($O \leq L$):
\begin{equation}
\label{eq:submap_def}
\begin{aligned}
\mathcal{S}_k &= \mathcal{O}_{k-1,k} \cup \{\, \mathbf{I}^{t,c} \,\}_{t=kL}^{(k+1)L-1}, \\[4pt]
\mathcal{O}_{k-1,k} &=
\begin{cases}
\{\, \mathbf{I}^{t,c} \,\}_{t=kL - O}^{kL - 1}, & k>0, \\[6pt]
\varnothing, & k=0.
\end{cases}
\end{aligned}
\end{equation}
Each submap is independently processed by a generic 3DFVM $\mathcal{M}$ to predict per-camera 3D attributes:
$$
\mathcal{M}(\mathcal{S}_k) = (\mathbf{K}^{t,c}, \mathbf{T}^{t,c}, \mathbf{P}^{t,c}),
$$
where $\mathbf{K}$, $\mathbf{T}$, and $\mathbf{P}$ denote the intrinsic matrix, camera pose, and reconstructed point cloud, respectively.
All poses and point clouds are expressed in the coordinate frame of the first camera within the submap.

Our goal is to align all submaps $\{\mathcal{S}_{k}\}_{k=0}^{K-1}$ into a globally consistent camera trajectory $\{\mathbf{T}^{t,c}\}_{t=0}^{T-1}$ and point cloud representation $\{\mathbf{P}^{t,c}\}_{t=0}^{T-1}$ through an online optimization process based solely on overlapping frames.
Our workflow is illustrated in Fig. \ref{fig:overview} and detailed in the following subsections:
Sect.~\ref{sec:pose} introduces the point-agnostic submap registration;
Sect.~\ref{sec:generation} presents the definition and generation of control points;
Sect.~\ref{sec:propatation} explains the temporal propagation of control points;
and Sect.~\ref{sec:tps} introduces the control point aggregation and the TPS deformation to align all submap point clouds into a globally consistent canonical space.


\subsection{Point-Agnostic Submap Registration}
\label{sec:pose}
Given two consecutive submaps $\mathcal{S}_{k-1}$ and $\mathcal{S}_k$ sharing overlapping frames $\mathcal{O}_{k-1,k}$,
we estimate a transformation $\mathbf{H}_{k \rightarrow k-1}$ that maps the coordinate frame of $\mathcal{S}_k$ to that of $\mathcal{S}_{k-1}$.
Previous approaches estimate $\mathbf{H}_{k \rightarrow k-1}$ as a $\mathrm{Sim}(3)$ \cite{vggtlong} or $\mathrm{SL}(4)$ \cite{vggtslam} transformation that best aligns the predicted point clouds of $\mathcal{O}_{k-1,k}$ in $\mathcal{S}_{k-1}$ and $\mathcal{S}_k$.
However, as analyzed in Sec.~\ref{sec:intro}, such point-based registrations not only compromise trajectory accuracy to offset local inconsistencies, but are also sensitive to geometry noise in predictions.

In contrast, we estimate $\mathbf{H}_{k \rightarrow k-1}$ directly from the overlapping camera poses, which are empirically more stable than raw point clouds.
Let $\{\mathbf{T}^{i}_{k-1}\}_{i=0}^{O-1}$ and $\{\mathbf{T}^{i}_{k}\}_{i=0}^{O-1}$ denote the predicted reference camera poses of $\mathcal{O}_{k-1,k}$ in $\mathcal{S}_{k-1}$ and $\mathcal{S}_k$.
Although these two pose sets describe the same $O$ cameras, 
they are expressed in distinct local coordinate frames.
By aligning them, we bring consecutive submaps into a unified global reference frame (that of $\mathcal{S}_{k-1}$).
Specifically, for each camera pair $(\mathbf{T}^{i}_{k-1}, \mathbf{T}^{i}_{k})$, 
we compute the inter-submap transformation as:
\begin{equation}
\mathbf{H}^{i}_{k \rightarrow k-1} = \mathbf{T}^{i}_{k-1} (\mathbf{T}^{i}_{k})^{-1},
\end{equation}
and obtain the final submap transformation $\mathbf{H}_{k \rightarrow k-1}$ by averaging $\{\mathbf{H}^{i}_{k \rightarrow k-1}\}_{i=0}^{O-1}$.
The rotational component is averaged using the Chordal $L_2$ rotation averaging method, which minimizes the sum of squared Frobenius norm of the difference between each rotation and the estimate ~\cite{Rotationaveraging}, while the translational components are averaged.

This minimalistic strategy empirically yields the most stable and accurate trajectory as shown in Sec. \ref{sec:exp_traj}.
Through the above procedure, all submaps are sequentially connected and transformed into the coordinate frame of the first image $\mathbf{I}_{0,0}$.
All poses and points mentioned hereafter are expressed in this unified reference frame.



\subsection{Control Point Definition and Generation}
\label{sec:generation}
We define a control point as a fixed spatial location in the underlying 3D world, which ideally remains invariant across all submaps. 
Formally, let $\mathbf{\bar{p}}^i$ denote the true world position of the $i$-th control point, and let $\{\mathbf{p}^i_k\}_{k=0}^{K-1}$ denote its predicted observations from $\mathcal{M}$ in the $K$ registered submaps. In a globally consistent reconstruction, all these observations should coincide, \emph{i.e.}, $\mathbf{p}^i_k = \mathbf{\bar{p}}^i$ for all $k \in \{0, \ldots, K-1\}$ whenever $\mathbf{p}^i_k$ exists.
However, independently inferred submaps inevitably exhibit geometric inconsistencies, causing the same physical location to be reconstructed as different 3D points across submaps. In other words, global control points and their per-submap observations encode the spatial-temporal distortions of geometry. 
Our goal is therefore to construct a set of global control points $\mathcal{C}_{\text{global}} = \{\mathbf{p}_k^i \mid k \in \{0, \ldots, K-1\}\}_{i=0}^{P-1}$ and their corresponding canonical locations $\mathcal{\bar{C}}_{\text{global}} = \{\mathbf{\bar{p}}^i\}_{i=0}^{P-1}$. By aligning $\mathcal{C}_{\text{global}}$ to $\mathcal{\bar{C}}_{\text{global}}$, we explicitly correct these distortions, yielding a globally consistent reconstruction.


To achieve even spatial coverage of control points throughout the 3D scene,
we first define a voxel-based control point generation function given a source point set 
$\{\mathbf{x}^i \in \mathbb{R}^3\}_{i=0}^{N-1}$ and a voxel size $\delta_v$:
\begin{equation}
\label{eq:voxelcontrol}
\mathcal{V}:\ (\{\mathbf{x}^i\}_{i=0}^{N-1},\, \delta_v)\ \mapsto\ \{\mathbf{p}^i\}_{i=0}^{P-1} \subset \{\mathbf{x}^i\}_{i=0}^{N-1},
\end{equation}
which voxelizes the input point cloud and selects one representative point per occupied voxel ($P\ll N$).
Specifically, we construct a uniform voxel grid in $\mathbb{R}^3$ with cell size $\delta_v$.  
Let $\mathbf{m} = \min_i \mathbf{x}^i$ denote the global minimum coordinate.
Each input point $\mathbf{x}^i$ is assigned to a voxel index
$\mathbf{v}^i = \Big\lfloor \frac{\mathbf{x}^i - \mathbf{m}}{\delta_v} \Big\rfloor \in \mathbb{Z}^3$, and the geometric center of voxel $\mathbf{v}$ is $\mathbf{c}(\mathbf{v}) = \mathbf{m} + \big(\mathbf{v} + \tfrac{1}{2}\big)\,\delta_v$.
Within each occupied voxel $\mathbf{v}$, we select the point closest to its center as its representative:
$i^\star(\mathbf{v}) = \underset{i:\,\mathbf{v}^i=\mathbf{v}}{\arg\min} 
\|\mathbf{x}^i - \mathbf{c}(\mathbf{v})\|_2^2$.
The final control point set is thus obtained as
$
\{\mathbf{p}^i\}_{i=0}^{P-1} = \{\, \mathbf{x}^{i^\star(\mathbf{v})} \mid \mathbf{v} \in \mathcal{V}_{\text{occ}} \,\}
$,
where $\mathcal{V}_{\text{occ}}$ denotes the set of occupied voxels.  
\subsection{Control Point Propagation}
\label{sec:propatation}

Given two consecutive submaps $\mathcal{S}_{k-1}$ and $\mathcal{S}_k$ sharing overlapping frames $\mathcal{O}_{k-1,k}$,
their predicted point clouds of $\mathcal{O}_{k-1,k}$ are denoted as $\mathcal{P}_{k-1}^{(k-1)\leftrightarrow k}$ and $\mathcal{P}_{k}^{(k-1)\leftrightarrow k}$, respectively.
We first extract a sparse set of control points from $\mathcal{P}_{k-1}^{(k-1)\leftrightarrow k}$ using the function defined in Eq.~\eqref{eq:voxelcontrol}:
\begin{equation}
\label{eq:newcontrol}
    \mathcal{C}_{k-1} = \{\mathbf{p}^i_{k-1}\}_{i=0}^{P_{k-1}-1} = \mathcal{V}(\mathcal{P}_{k-1}^{(k-1)\leftrightarrow k},\, \delta_v).
\end{equation}
For clarity, we describe the subsequent process using a single control point $\mathbf{p}_{k-1} \in \mathcal{C}_{k-1}$ and omit the superscript $i$.
Owing to the pixel-aligned nature of foundation models, each 3D point $\mathbf{p}_{k-1} \in \mathcal{P}_{k-1}^{(k-1)\leftrightarrow k}$ inherently originates from a pixel ${\mathbf{u}}_{k-1}$ in an overlapping image $\mathbf{I} \in \mathcal{O}_{k-1,k}$.
Since $\mathbf{I} \in \mathcal{O}_{k-1,k} \subset \mathcal{S}_k$, we can unproject the same pixel ${\mathbf{u}}_{k-1}$ into a 3D position $\mathbf{p}_{k} \in \mathcal{P}_{k}^{(k-1)\leftrightarrow k}$ using the camera parameters of $\mathcal{S}_{k}$.
This establishes a control-point correspondence between two consecutive submaps representing the same physical location bridged by overlapping frames. 

To achieve global consistency beyond short-term optimality, we further propagate control points across the entire sequence.
Specifically, we illustrate how to propagate the control point $\mathbf{p}_k$ of $\mathcal{S}_{k}$, originally generated from $\mathcal{S}_{k-1}$, to the next submap $\mathcal{S}_{k+1}$, which shares the overlapping frames $\mathcal{O}_{k,k+1}$ with $\mathcal{S}_k$.
We first project $\mathbf{p}_k \in \mathcal{P}_{k}^{(k-1)\leftrightarrow k}$ onto an image in $\mathcal{O}_{k,k+1}$, obtaining its pixel location ${\mathbf{u}}_{k}$.
In practice, it may project onto multiple images within $\mathcal{O}_{k,k+1}$; we retain the one with the smallest reprojection error.
Since $\mathcal{O}_{k,k+1} \subset \mathcal{S}_{k+1}$, we unproject the same pixel ${\mathbf{u}}_{k+1} = {\mathbf{u}}_{k}$ into 3D position $\mathbf{p}_{k+1} \in \mathcal{P}_{k+1}^{k\leftrightarrow k+1}$.
Repeating this process forms a rich observation sequence for the same world location.
If a control point fails to obtain a valid projection in any submap, its propagation is terminated.




In implementation, when performing voxelization for new control-point generation (Eq.~\eqref{eq:newcontrol}) in a new submap $\mathcal{S}_k$,
control points from the previous submap $\mathcal{S}_{k-1}$ are first propagated forward to $\mathcal{S}_k$.
A new control point is instantiated only if its voxel is not already occupied by any propagated control point from $\mathcal{S}_{k-1}$.
Conversely, each newly created control point is back-propagated to $\mathcal{S}_{k-1}$ to enrich mutual observations between the two submaps.
All control point observations are maintained in a global pool
$\mathcal{C}_{\text{global}} = \{\mathbf{p}_k^i \mid k \in \{0,\ldots,K-1\}\}_{i=0}^{P-1}$,
forming a globally connected control-point graph that serves as the foundation for the subsequent TPS alignment.





\begin{table*}[t]
    \centering
    \vspace{-2ex}
    \caption{
        Camera trajectory accuracy on \textbf{Waymo} \cite{waymo}, reported as ATE (RMSE [m]), RTE (RMSE [m]), and RRE (RMSE [$^\circ$]).
        Best results under each foundation model are \textbf{\underline{bold underlined}}, and \textcolor{failure}{catastrophic failures} (\text{ATE RMSE} $>$ 5\% of GT trajectory length) are in red.
    }\vspace{-2ex}
    \tablestyle{2pt}{1.05}
    \resizebox{0.99\linewidth}!{
    \begin{tabular}{ccccccccccccccccccccccccccccc}
        \toprule[0.17em]
        \multirow{3}{*}{\textbf{\makecell{Foundation\\Model}}} &
        \multirow{3}{*}{\textbf{\makecell{Alignment\\Strategy}}} &
        \multicolumn{3}{c}{\textbf{163453191685903.}} & 
        \multicolumn{3}{c}{\textbf{183829460855609.}} & 
        \multicolumn{3}{c}{\textbf{315615587265462.}} &
        \multicolumn{3}{c}{\textbf{346181117917711.}} & 
        \multicolumn{3}{c}{\textbf{405841035328651.}} &
        \multicolumn{3}{c}{\textbf{520018670674820.}} & 
        \multicolumn{3}{c}{\textbf{610454533463565.}} &
        \multicolumn{3}{c}{\textbf{Avg.}} \\
        \cmidrule(lr){3-5}
        \cmidrule(lr){6-8}
        \cmidrule(lr){9-11}
        \cmidrule(lr){12-14}
        \cmidrule(lr){15-17}
        \cmidrule(lr){18-20}
        \cmidrule(lr){21-23}
        \cmidrule(lr){24-26}
        & &
        ATE$^\downarrow$ & RTE$^\downarrow$ & RRE$^\downarrow$ &
        ATE$^\downarrow$ & RTE$^\downarrow$ & RRE$^\downarrow$ &
        ATE$^\downarrow$ & RTE$^\downarrow$ & RRE$^\downarrow$ &
        ATE$^\downarrow$ & RTE$^\downarrow$ & RRE$^\downarrow$ &
        ATE$^\downarrow$ & RTE$^\downarrow$ & RRE$^\downarrow$ &
        ATE$^\downarrow$ & RTE$^\downarrow$ & RRE$^\downarrow$ &
        ATE$^\downarrow$ & RTE$^\downarrow$ & RRE$^\downarrow$ &
        ATE$^\downarrow$ & RTE$^\downarrow$ & RRE$^\downarrow$ \\
        \midrule[0.08em]

        \multirow{3}{*}{VGGT \cite{vggt}}
        & VGGT-Long \cite{vggtlong}
        & 1.85 & 0.36 & 0.54 & 0.28 & 0.20 & 0.50 & 1.59 & 0.38 & 0.78 & \textbf{\underline{1.45}} & \textbf{\underline{0.40}} & 0.89 & 1.41 & 0.31 & 0.66 & 2.76 & \textbf{\underline{0.38}} & 1.07 & 0.55 & 0.24 & 0.56 & 1.42 & 0.32 & 0.71 \\
        & VGGT-SLAM \cite{vggtslam}
        & \textcolor{failure}{12.45} & \textcolor{failure}{4.16} & \textcolor{failure}{4.39} & 1.04 & 0.97 & 3.84 & \textcolor{failure}{13.18} & \textcolor{failure}{4.01} & \textcolor{failure}{4.79} & \textcolor{failure}{16.35} & \textcolor{failure}{11.16} & \textcolor{failure}{17.32} & \textcolor{failure}{7.04} & \textcolor{failure}{3.88} & \textcolor{failure}{10.36} & \textcolor{failure}{34.75} & \textcolor{failure}{13.72} & \textcolor{failure}{31.05} & 0.64 & 0.60 & 4.58 & 12.21 & 5.50 & 10.90 \\
        &\cellcolor{blue!8} TALO
        &\cellcolor{blue!8} \textbf{\underline{1.37}} &\cellcolor{blue!8} \textbf{\underline{0.32}} &\cellcolor{blue!8} \textbf{\underline{0.15}} &\cellcolor{blue!8} \textbf{\underline{0.10}} &\cellcolor{blue!8} \textbf{\underline{0.07}} &\cellcolor{blue!8} \textbf{\underline{0.08}} &\cellcolor{blue!8} \textbf{\underline{0.63}} &\cellcolor{blue!8} \textbf{\underline{0.23}} &\cellcolor{blue!8} \textbf{\underline{0.13}} &\cellcolor{blue!8} 2.30 &\cellcolor{blue!8} 0.68 &\cellcolor{blue!8} \textbf{\underline{0.14}} &\cellcolor{blue!8} \textbf{\underline{0.91}} &\cellcolor{blue!8} \textbf{\underline{0.19}} &\cellcolor{blue!8} \textbf{\underline{0.13}} &\cellcolor{blue!8} \textbf{\underline{2.21}} &\cellcolor{blue!8} 0.40 &\cellcolor{blue!8} \textbf{\underline{0.25}} &\cellcolor{blue!8} \textbf{\underline{0.10}} &\cellcolor{blue!8} \textbf{\underline{0.07}} &\cellcolor{blue!8} \textbf{\underline{0.08}} &\cellcolor{blue!8} \textbf{\underline{1.09}} &\cellcolor{blue!8} \textbf{\underline{0.28}} &\cellcolor{blue!8} \textbf{\underline{0.14}} \\
        \midrule[0.08em]

        \multirow{3}{*}{$\pi^3$ \cite{pi3}}
        & VGGT-Long \cite{vggtlong}
        & \textbf{\underline{0.95}} & 0.30 & 0.71 & 0.61 & 0.29 & 0.84 & 1.69 & 0.71 & 1.25 & 7.91 & 0.96 & 1.06 & 1.05 & 0.33 & 0.81 & 2.98 & 0.48 & 1.37 & 0.37 & 0.26 & 0.48 & 2.22 & 0.48 & 0.93 \\
        & VGGT-SLAM \cite{vggtslam}
        & \textcolor{failure}{19.56} & \textcolor{failure}{6.52} & \textcolor{failure}{4.78} & 1.03 & 0.59 & 4.47 & \textcolor{failure}{29.85} & \textcolor{failure}{9.49} & \textcolor{failure}{6.87} & \textcolor{failure}{65.73} & \textcolor{failure}{13.29} & \textcolor{failure}{8.58} & 2.00 & 1.98 & 7.42 & \textcolor{failure}{35.84} & \textcolor{failure}{6.94} & \textcolor{failure}{30.46} & 1.59 & 0.67 & 6.16 & 22.23 & 5.64 & 9.82 \\
        &\cellcolor{blue!8} TALO
        &\cellcolor{blue!8} 0.97 &\cellcolor{blue!8} \textbf{\underline{0.23}} &\cellcolor{blue!8} \textbf{\underline{0.31}} &\cellcolor{blue!8} \textbf{\underline{0.15}} &\cellcolor{blue!8} \textbf{\underline{0.10}} &\cellcolor{blue!8} \textbf{\underline{0.22}} &\cellcolor{blue!8} \textbf{\underline{1.29}} &\cellcolor{blue!8} \textbf{\underline{0.40}} &\cellcolor{blue!8} \textbf{\underline{0.28}} &\cellcolor{blue!8} \textbf{\underline{2.02}} &\cellcolor{blue!8} \textbf{\underline{0.52}} &\cellcolor{blue!8} \textbf{\underline{0.17}} &\cellcolor{blue!8} \textbf{\underline{0.24}} &\cellcolor{blue!8} \textbf{\underline{0.09}} &\cellcolor{blue!8} \textbf{\underline{0.15}} &\cellcolor{blue!8} \textbf{\underline{1.03}} &\cellcolor{blue!8} \textbf{\underline{0.36}} &\cellcolor{blue!8} \textbf{\underline{0.45}} &\cellcolor{blue!8} \textbf{\underline{0.31}} &\cellcolor{blue!8} \textbf{\underline{0.14}} &\cellcolor{blue!8} \textbf{\underline{0.10}} &\cellcolor{blue!8} \textbf{\underline{0.86}} &\cellcolor{blue!8} \textbf{\underline{0.26}} &\cellcolor{blue!8} \textbf{\underline{0.24}} \\

          \midrule[0.08em]
        \multirow{3}{*}{Map. \cite{mapanything}} 
        & VGGT-Long~\cite{vggtlong}  
        & 2.71 & 0.37 & 1.61 & 0.40 & 0.21 & 0.43 & \textcolor{failure}{10.60} & \textcolor{failure}{0.99} & \textcolor{failure}{3.34} & 7.26 & 0.87 & 1.33 & 1.16 & 0.33 & 1.17 & \textbf{\underline{3.05}} & 1.30 & 2.89 & 0.61 & 0.36 & 1.19 & 3.68 & 0.63 & 1.71 \\
        
        & VGGT-SLAM~\cite{vggtslam}  
        & \textcolor{failure}{46.17} & \textcolor{failure}{7.09} & \textcolor{failure}{29.35} & 4.62 & 2.93 & 6.03 & \textcolor{failure}{48.03} & \textcolor{failure}{5.27} & \textcolor{failure}{31.08} & \textcolor{failure}{96.80} & \textcolor{failure}{50.48} & \textcolor{failure}{41.68} & \textcolor{failure}{6.25} & \textcolor{failure}{2.87} & \textcolor{failure}{14.36} & \textcolor{failure}{10.46} & \textcolor{failure}{8.35} & \textcolor{failure}{38.96} & 1.16 & 1.22 & 3.56 & 30.50 & 11.17 & 23.57 \\
        & \cellcolor{blue!8}TALO 
        & 
         \cellcolor{blue!8}\textbf{\underline{1.55}} & \cellcolor{blue!8} \textbf{\underline{0.28}} &\cellcolor{blue!8} \textbf{\underline{0.96}} &\cellcolor{blue!8} \textbf{\underline{0.21}} &\cellcolor{blue!8} \textbf{\underline{0.12}} &\cellcolor{blue!8} \textbf{\underline{0.15}} &\cellcolor{blue!8} \textbf{\underline{0.72}} &\cellcolor{blue!8} \textbf{\underline{0.46}} &\cellcolor{blue!8} \textbf{\underline{0.27}} &\cellcolor{blue!8} \textbf{\underline{1.62}} &\cellcolor{blue!8} \textbf{\underline{0.64}} &\cellcolor{blue!8} \textbf{\underline{0.26}} &\cellcolor{blue!8} \textbf{\underline{0.56}} &\cellcolor{blue!8} \textbf{\underline{0.27}} &\cellcolor{blue!8} \textbf{\underline{0.52}} &\cellcolor{blue!8} 5.00 &\cellcolor{blue!8} \textbf{\underline{1.01}} &\cellcolor{blue!8} \textbf{\underline{1.86}} &\cellcolor{blue!8} \textbf{\underline{0.18}} &\cellcolor{blue!8} \textbf{\underline{0.19}} &\cellcolor{blue!8} \textbf{\underline{0.20}} &\cellcolor{blue!8} \textbf{\underline{1.40}} &\cellcolor{blue!8} \textbf{\underline{0.42}} &\cellcolor{blue!8} \textbf{\underline{0.60}} \\
        
        \bottomrule[0.17em]
    \end{tabular}
    }
    
    \label{tab:waymo_traj}
\end{table*}

\begin{table*}[t]
    \centering
    \caption{
        Camera trajectory accuracy on \textbf{nuScenes} \cite{nuScenes}, reported as ATE (RMSE [m]), RTE (RMSE [m]), and RRE (RMSE [$^\circ$]).
        Best results under each foundation model are \textbf{\underline{bold underlined}}, and \textcolor{failure}{catastrophic failures} (\text{ATE RMSE} $>$ 5\% of GT trajectory length) are in red.
    }\vspace{-2ex}
    \tablestyle{2pt}{1.05}
    \resizebox{0.99\linewidth}!{
    \begin{tabular}{cccccccccccccccccccccccccc}
        \toprule[0.17em]
        \multirow{3}{*}{\textbf{\makecell{Foundation\\Model}}} &
        \multirow{3}{*}{\textbf{\makecell{Alignment\\Strategy}}} &
        \multicolumn{3}{c}{\textbf{scene-0003} } & 
        \multicolumn{3}{c}{\textbf{scene-0012} } & 
        \multicolumn{3}{c}{\textbf{scene-0013} } & 
        \multicolumn{3}{c}{\textbf{scene-0036} } & 
        \multicolumn{3}{c}{\textbf{scene-0039} } & 
        \multicolumn{3}{c}{\textbf{scene-0092} } & 
        \multicolumn{3}{c}{\textbf{scene-0094} } & 
        \multicolumn{3}{c}{\textbf{Avg.}} \\
        \cmidrule(lr){3-5}
        \cmidrule(lr){6-8}
        \cmidrule(lr){9-11}
        \cmidrule(lr){12-14}
        \cmidrule(lr){15-17}
        \cmidrule(lr){18-20}
        \cmidrule(lr){21-23}
        \cmidrule(lr){24-26}
        & &
        ATE$^\downarrow$ & RTE$^\downarrow$ & RRE$^\downarrow$ &
        ATE$^\downarrow$ & RTE$^\downarrow$ & RRE$^\downarrow$ &
        ATE$^\downarrow$ & RTE$^\downarrow$ & RRE$^\downarrow$ &
        ATE$^\downarrow$ & RTE$^\downarrow$ & RRE$^\downarrow$ &
        ATE$^\downarrow$ & RTE$^\downarrow$ & RRE$^\downarrow$ &
        ATE$^\downarrow$ & RTE$^\downarrow$ & RRE$^\downarrow$ &
        ATE$^\downarrow$ & RTE$^\downarrow$ & RRE$^\downarrow$ &
        ATE$^\downarrow$ & RTE$^\downarrow$ & RRE$^\downarrow$ \\
        \midrule[0.08em]

        \multirow{3}{*}{VGGT \cite{vggt}} 
        & VGGT-Long~\cite{vggtlong} 
        & 0.31 & 0.18 & 0.40 & 1.61 & 0.48 & 0.80 & 1.37 & 0.43 & 0.63 & 2.52 & 0.63 & 0.62 & \textbf{\underline{1.45}} & \textbf{\underline{0.44}} & 0.70 & 3.02 & 0.65 & 0.44 & \textbf{\underline{1.15}} & 0.48 & 0.47 & 1.63 & 0.47 & 0.58 \\
        & VGGT-SLAM~\cite{vggtslam}  
        & \textbf{\underline{0.19}} & 0.24 & 1.57 & \textcolor{failure}{19.58} & \textcolor{failure}{4.24} & \textcolor{failure}{3.48} & \textcolor{failure}{11.58} & \textcolor{failure}{2.74} & \textcolor{failure}{5.56} & \textcolor{failure}{21.09} & \textcolor{failure}{4.66} & \textcolor{failure}{3.35} & \textcolor{failure}{28.42} & \textcolor{failure}{4.73} & \textcolor{failure}{1.67} & \textcolor{failure}{11.02} & \textcolor{failure}{2.06} & \textcolor{failure}{0.98} & \textcolor{failure}{30.83} & \textcolor{failure}{4.08} & \textcolor{failure}{28.93} & 17.53 & 3.25 & 6.51 \\
        &\cellcolor{blue!8} TALO 
        &\cellcolor{blue!8} 0.36 &\cellcolor{blue!8} \textbf{\underline{0.13}} &\cellcolor{blue!8} \textbf{\underline{0.20}} &\cellcolor{blue!8} \textbf{\underline{1.11}} &\cellcolor{blue!8} \textbf{\underline{0.40}} &\cellcolor{blue!8} \textbf{\underline{0.35}} &\cellcolor{blue!8} \textbf{\underline{0.71}} &\cellcolor{blue!8} \textbf{\underline{0.31}} &\cellcolor{blue!8} \textbf{\underline{0.17}} &\cellcolor{blue!8} \textbf{\underline{1.05}} &\cellcolor{blue!8} \textbf{\underline{0.40}} &\cellcolor{blue!8} \textbf{\underline{0.18}} &\cellcolor{blue!8} 1.75 &\cellcolor{blue!8} 0.48 &\cellcolor{blue!8} \textbf{\underline{0.20}} &\cellcolor{blue!8} \textbf{\underline{3.01}} &\cellcolor{blue!8} \textbf{\underline{0.61}} &\cellcolor{blue!8} \textbf{\underline{0.14}} &\cellcolor{blue!8} 1.17 &\cellcolor{blue!8} \textbf{\underline{0.29}} &\cellcolor{blue!8} \textbf{\underline{0.11}} &\cellcolor{blue!8} \textbf{\underline{1.31}} &\cellcolor{blue!8} \textbf{\underline{0.37}} &\cellcolor{blue!8} \textbf{\underline{0.19}} \\
        \midrule[0.08em]

        \multirow{3}{*}{$\pi^3$ \cite{pi3}} 
        & VGGT-Long~\cite{vggtlong}  
        &0.49 & 0.31 & 0.99 & 2.26 & 0.66 & 1.40 & \textbf{\underline{1.60}} & 1.07 & 4.47 & 2.20 & 0.42 & 0.73 & 1.57 & \textbf{\underline{0.32}} & 0.97 & 1.17 & 0.49 & 0.75 & 2.16 & 0.96 & 1.09 & 1.63 & 0.60 & 1.49 \\
        & VGGT-SLAM~\cite{vggtslam}  
        & 0.42 & 0.37 & 2.46 & 8.14 & 6.34 & 12.75 & \textcolor{failure}{12.85} & \textcolor{failure}{3.83} & \textcolor{failure}{11.25} & 7.89 & 8.02 & 12.19 & \textcolor{failure}{25.37} & \textcolor{failure}{6.38} & \textcolor{failure}{10.04} & 2.52 & 4.01 & 4.99 & \textcolor{failure}{8.39} & \textcolor{failure}{2.46} & \textcolor{failure}{1.86} & 9.37 & 4.49 & 7.93 \\
        &\cellcolor{blue!8} TALO 
        &\cellcolor{blue!8} \textbf{\underline{0.31}} &\cellcolor{blue!8} \textbf{\underline{0.17}} &\cellcolor{blue!8} \textbf{\underline{0.34}} &\cellcolor{blue!8} \textbf{\underline{1.08}} &\cellcolor{blue!8} \textbf{\underline{0.43}} &\cellcolor{blue!8} \textbf{\underline{0.46}} &\cellcolor{blue!8} 1.85 &\cellcolor{blue!8} \textbf{\underline{0.86}} &\cellcolor{blue!8} \textbf{\underline{0.78}} &\cellcolor{blue!8} \textbf{\underline{1.34}} &\cellcolor{blue!8} \textbf{\underline{0.39}} &\cellcolor{blue!8} \textbf{\underline{0.24}} &\cellcolor{blue!8} \textbf{\underline{1.30}} &\cellcolor{blue!8} 0.36 &\cellcolor{blue!8} \textbf{\underline{0.19}} &\cellcolor{blue!8} \textbf{\underline{0.61}} &\cellcolor{blue!8} \textbf{\underline{0.32}} &\cellcolor{blue!8} \textbf{\underline{0.35}} &\cellcolor{blue!8} \textbf{\underline{0.63}} &\cellcolor{blue!8} \textbf{\underline{0.32}} &\cellcolor{blue!8} \textbf{\underline{0.30}} &\cellcolor{blue!8} \textbf{\underline{1.02}} &\cellcolor{blue!8} \textbf{\underline{0.41}} &\cellcolor{blue!8} \textbf{\underline{0.38}} \\

        \midrule[0.08em]
        \multirow{3}{*}{Map. \cite{mapanything}} 
        & VGGT-Long~\cite{vggtlong}  
        & 0.34 & 0.21 & 0.43 & 1.76 & 0.48 & 1.25 & 1.79 & 0.38 & 1.35 & 4.58 & 0.52 & 1.48 & 3.36 & 0.68 & 1.97 & 1.72 & 1.10 & 1.23 & 2.80 & 0.61 & 1.64 & 2.34 & 0.57 & 1.34 \\
        & VGGT-SLAM~\cite{vggtslam}  
        & 1.12 & 0.46 & 6.71 & \textcolor{failure}{53.69} & \textcolor{failure}{6.45} & \textcolor{failure}{33.69} & \textcolor{failure}{21.76} & \textcolor{failure}{3.74} & \textcolor{failure}{11.96} & \textcolor{failure}{38.99} & \textcolor{failure}{12.24} & \textcolor{failure}{42.71} & \textcolor{failure}{50.34} & \textcolor{failure}{24.39} & \textcolor{failure}{19.34} & 6.81 & 2.51 & 5.24 & 4.31 & 1.98 & 5.25 & 25.29 & 7.40 & 17.84 \\
        &\cellcolor{blue!8} TALO 
        &\cellcolor{blue!8} \textbf{\underline{0.21}} &\cellcolor{blue!8} \textbf{\underline{0.14}} &\cellcolor{blue!8} \textbf{\underline{0.23}} &\cellcolor{blue!8} \textbf{\underline{0.75}} &\cellcolor{blue!8} \textbf{\underline{0.36}} &\cellcolor{blue!8} \textbf{\underline{0.44}} &\cellcolor{blue!8} \textbf{\underline{0.46}} &\cellcolor{blue!8} \textbf{\underline{0.33}} &\cellcolor{blue!8} \textbf{\underline{0.28}} &\cellcolor{blue!8} \textbf{\underline{1.25}} &\cellcolor{blue!8} \textbf{\underline{0.51}} &\cellcolor{blue!8} \textbf{\underline{0.28}} &\cellcolor{blue!8} \textbf{\underline{1.91}} &\cellcolor{blue!8} \textbf{\underline{0.61}} &\cellcolor{blue!8} \textbf{\underline{0.28}} &\cellcolor{blue!8} \textbf{\underline{0.89}} &\cellcolor{blue!8} \textbf{\underline{0.62}} &\cellcolor{blue!8} \textbf{\underline{0.26}} &\cellcolor{blue!8} \textbf{\underline{0.89}} &\cellcolor{blue!8} \textbf{\underline{0.40}} &\cellcolor{blue!8} \textbf{\underline{0.19}} &\cellcolor{blue!8} \textbf{\underline{0.91}} &\cellcolor{blue!8} \textbf{\underline{0.42}} &\cellcolor{blue!8} \textbf{\underline{0.28}} \\

        \bottomrule[0.17em]

    \end{tabular}\vspace{-2ex}
    }
    \label{tab:nuscenes_traj}
\end{table*}

\subsection{Thin Plate Spline Deformation}
\label{sec:tps}
Given the global control point observations 
$\mathcal{C}_{\text{global}} = \{\mathbf{p}_k^i \mid k \in \{0,\dots,K-1\}\}_{i=0}^{P-1}$,
we first estimate the canonical 3D position of the $i$-th control point by robustly aggregating its observations across multiple submaps:
\begin{equation}
\mathbf{\bar{p}}^i =
\mathrm{Aggregate}\big(\{\mathbf{p}_k^i \mid k \in \{0,\dots,K-1\}\}\big),
\end{equation}
where the aggregation is implemented as a MAD-filtered mean to suppress the influence of dynamic objects and outliers.
Let $\mathcal{C}_{k} = \{\mathbf{p}_k^i\}_{i=0}^{P_k-1}$ denote all control points observed in $\mathcal{S}_k$, and 
$\mathcal{\bar{C}}_{k} = \{\mathbf{\bar{p}}^i\}_{i=0}^{P_k-1}$ their corresponding canonical positions.
Before TPS fitting, we further apply a local Gaussian smoothing to the canonical targets for each submap.
Specifically, for each point $\mathbf{p}_k^i \in \mathcal{C}_{k}$, we find its $Q$ nearest neighbors $\mathcal{N}_Q(i)$ based on Euclidean distance
$d_{ij} = \|\mathbf{p}_k^i - \mathbf{p}_k^j\|_2$, and compute a Gaussian-weighted average of neighbor displacements:
\[
\tilde{w}_{ij} = \exp\!\Big(-\tfrac{d_{ij}^2}{2\sigma^2}\Big),
\qquad
w_{ij} =
\frac{\tilde{w}_{ij}}
{\sum_{l \in \mathcal{N}_Q(i)} \tilde{w}_{il}}.
\]
Its smoothed canonical target for submap $k$ is given by:
\[
\bar{\mathbf{p}}_{k}^{i,\,\text{sm}}
=
\mathbf{p}_k^i
+
\sum_{j \in \mathcal{N}_Q(i)} w_{ij}\,(\bar{\mathbf{p}}^{j} - \mathbf{p}_k^j).
\]
The bandwidth $\sigma$ is automatically estimated as the median distance to the $M$-th nearest neighbor, ensuring scale adaptivity to local point density.
This Gaussian filtering enforces local spatial coherence in both the magnitude and direction of corrections, 
effectively reducing noise and stabilizing the subsequent TPS deformation.

We then employ a 3D Thin Plate Spline (TPS) \cite{tps1,tps2,tps3,tps4} deformation $\mathcal{F}_k:\mathbb{R}^3 \rightarrow \mathbb{R}^3$, 
a classical kernel-based deformation model that minimizes bending energy while interpolating given correspondences.
TPS provides a continuous spatial transformation that preserves global smoothness and local rigidity, making it well-suited for correcting local geometric distortions in our setting.
The TPS deformation of $\mathcal{S}_k$ is parameterized as:
\begin{equation}
\mathcal{F}_k(\mathbf{x}) = A_k\,\mathbf{x} + \mathbf{b}_k
+ \sum_{i=0}^{P_k-1} \mathbf{w}_k^i\,\phi(\|\mathbf{x}-\mathbf{p}_k^i\|),
\end{equation}
where $\mathbf{x} \in \mathbb{R}^3$ denotes any spatial position, 
$A_k \in \mathbb{R}^{3\times3}$ and $\mathbf{b}_k \in \mathbb{R}^3$ model the global affine component,
$\mathbf{w}_k^i \in \mathbb{R}^{3}$ are the TPS kernel weights associated with control points $\{\mathbf{p}_k^i\}$,
and $\phi(r)$ is a radial basis kernel that defines the non-rigid influence of each $\mathbf{p}_k^i$, which is set to $\phi(r)=r$ following the standard biharmonic radial basis in 3D.
The parameters $\{A_k, \mathbf{b}_k, \mathbf{w}_k^i\}$ are estimated by minimizing the energy:
\begin{equation}
\min_{A_k, \mathbf{b}_k, \mathbf{w}_k}
\sum_{i=0}^{P_k-1} 
\big\|\mathcal{F}_k(\mathbf{p}_k^i) - \bar{\mathbf{p}}_{k}^{i,\text{sm}}\big\|^2
+ \lambda\,\mathrm{tr}(\mathbf{w}_k^\top K_k \mathbf{w}_k),
\end{equation}
where $(K_k)_{ij} = \phi(\|\mathbf{p}_k^i - \mathbf{p}_k^j\|)$ encodes pairwise radial interactions among control points,
and $\lambda$ controls the smoothness regularization (bending energy).

This closed-form solution yields a smooth, differentiable deformation $\mathcal{F}_k$
that aligns the submap $\mathcal{S}_k$ to the global geometric consensus
by interpolating the control-point correspondences and smoothly deforming the remaining points.
Applying $\mathcal{F}_k$ to all points within $\mathcal{S}_k$ produces a locally consistent, globally coherent alignment across all submaps.
Importantly, TPS alignment can be applied at \textit{arbitrary} intervals using the currently accumulated control points to provide timely updates.
This \textit{proactive} correction complements traditional \textit{passive} mechanisms that rely on external signals such as loop closure detections.

\begin{table*}[t]
    \centering\vspace{-2ex}
    \caption{
        Reconstructed geometry evaluation on the \textbf{Waymo} \cite{waymo} dataset, reported as Accuracy, Completeness, and Chamfer Distance.
    }\vspace{-2ex}
    \tablestyle{2pt}{1.05}
    \resizebox{0.99\linewidth}!{
    \begin{tabular}{ccccccccccccccccccccccccccccc}
        \toprule[0.17em]
        \multirow{3}{*}{\textbf{\makecell{Foundation\\Model}}} &
        \multirow{3}{*}{\textbf{\makecell{Alignment\\Strategy}}} &
        \multicolumn{3}{c}{\textbf{163453191685903.}} & 
        \multicolumn{3}{c}{\textbf{183829460855609.}} & 
        \multicolumn{3}{c}{\textbf{315615587265462.}} &
        \multicolumn{3}{c}{\textbf{346181117917711.}} & 
        \multicolumn{3}{c}{\textbf{405841035328651.}} &
        \multicolumn{3}{c}{\textbf{520018670674820.}} & 
        \multicolumn{3}{c}{\textbf{610454533463565.}} &
        \multicolumn{3}{c}{\textbf{Avg.}} \\
        \cmidrule(lr){3-5}
        \cmidrule(lr){6-8}
        \cmidrule(lr){9-11}
        \cmidrule(lr){12-14}
        \cmidrule(lr){15-17}
        \cmidrule(lr){18-20}
        \cmidrule(lr){21-23}
        \cmidrule(lr){24-26}
        & &
        Acc.$^\downarrow$ & Com.$^\downarrow$ & Cha.$^\downarrow$ &
        Acc.$^\downarrow$ & Com.$^\downarrow$ & Cha.$^\downarrow$ &
        Acc.$^\downarrow$ & Com.$^\downarrow$ & Cha.$^\downarrow$ &
        Acc.$^\downarrow$ & Com.$^\downarrow$ & Cha.$^\downarrow$ &
        Acc.$^\downarrow$ & Com.$^\downarrow$ & Cha.$^\downarrow$ &
        Acc.$^\downarrow$ & Com.$^\downarrow$ & Cha.$^\downarrow$ &
        Acc.$^\downarrow$ & Com.$^\downarrow$ & Cha.$^\downarrow$ &
        Acc.$^\downarrow$ & Com.$^\downarrow$ & Cha.$^\downarrow$ 
        \\
        \midrule[0.08em]

        \multirow{3}{*}{VGGT \cite{vggt}} 
        & VGGT-Long~\cite{vggtlong} 
        &0.62 & 0.84 & 0.73 & 0.26 & \textbf{\underline{0.71}} & \textbf{\underline{0.49}} & 0.76 & 0.78 & 0.77 & 0.85 & 1.40 & 1.13 & 0.55 & 0.80 & 0.68 & 0.89 & \textbf{\underline{0.99}} & 0.94 & 0.56 & 0.48 & 0.52 & 0.64 & 0.86 & 0.75 \\
        & VGGT-SLAM~\cite{vggtslam}  
        & 3.87 & 6.50 & 5.18 & 5.04 & 6.44 & 5.74 & 9.06 & 8.67 & 8.87 & 9.46 & 9.39 & 9.43 & 7.37 & 8.24 & 7.80 & 5.22 & 9.24 & 7.23 & 5.12 & 3.66 & 4.39 & 6.45 & 7.45 & 6.95 \\
        &\cellcolor{blue!8} TALO 
        &\cellcolor{blue!8} \textbf{\underline{0.46}} &\cellcolor{blue!8} \textbf{\underline{0.58}} &\cellcolor{blue!8} \textbf{\underline{0.52}} &\cellcolor{blue!8} \textbf{\underline{0.18}} &\cellcolor{blue!8} 0.82 &\cellcolor{blue!8} 0.50 &\cellcolor{blue!8} \textbf{\underline{0.48}} &\cellcolor{blue!8} \textbf{\underline{0.63}} &\cellcolor{blue!8} \textbf{\underline{0.56}} &\cellcolor{blue!8} \textbf{\underline{0.61}} &\cellcolor{blue!8} \textbf{\underline{1.28}} &\cellcolor{blue!8} \textbf{\underline{0.94}} &\cellcolor{blue!8} \textbf{\underline{0.37}} &\cellcolor{blue!8} \textbf{\underline{0.68}} &\cellcolor{blue!8} \textbf{\underline{0.52}} &\cellcolor{blue!8} \textbf{\underline{0.74}} &\cellcolor{blue!8} 1.06 &\cellcolor{blue!8} \textbf{\underline{0.90}} &\cellcolor{blue!8} \textbf{\underline{0.35}} &\cellcolor{blue!8} \textbf{\underline{0.40}} &\cellcolor{blue!8} \textbf{\underline{0.37}} &\cellcolor{blue!8} \textbf{\underline{0.45}} &\cellcolor{blue!8} \textbf{\underline{0.78}} &\cellcolor{blue!8} \textbf{\underline{0.62}} \\
        
        \midrule[0.08em]

        \multirow{3}{*}{$\pi^3$ \cite{pi3}} 
        & VGGT-Long~\cite{vggtlong}  
        & 0.39 & \textbf{\underline{0.50}} & 0.45 & 0.41 & 0.67 & 0.54 & 0.86 & 1.15 & 1.00 & 1.32 & 2.00 & 1.66 & 0.41 & 0.53 & 0.47 & 1.15 & 1.11 & 1.13 & 0.49 & 0.34 & 0.42 & 0.72 & 0.90 & 0.81 \\
        & VGGT-SLAM~\cite{vggtslam}  
        & 8.51 & 9.14 & 8.82 & 2.77 & 3.24 & 3.01 & 8.63 & 9.49 & 9.06 & 9.53 & 9.63 & 9.58 & 3.07 & 3.41 & 3.24 & 1.68 & 7.79 & 4.73 & 0.96 & 1.16 & 1.06 & 5.02 & 6.27 & 5.64 \\
        &\cellcolor{blue!8} TALO 
        &\cellcolor{blue!8} \textbf{\underline{0.37}} &\cellcolor{blue!8} \textbf{\underline{0.50}} &\cellcolor{blue!8} \textbf{\underline{0.43}} &\cellcolor{blue!8} \textbf{\underline{0.18}} &\cellcolor{blue!8} \textbf{\underline{0.56}} &\cellcolor{blue!8} \textbf{\underline{0.37}} &\cellcolor{blue!8} \textbf{\underline{0.67}} &\cellcolor{blue!8} \textbf{\underline{0.87}} &\cellcolor{blue!8} \textbf{\underline{0.77}} &\cellcolor{blue!8} \textbf{\underline{1.21}} &\cellcolor{blue!8} \textbf{\underline{1.42}} &\cellcolor{blue!8} \textbf{\underline{1.31}} &\cellcolor{blue!8} \textbf{\underline{0.18}} &\cellcolor{blue!8} \textbf{\underline{0.41}} &\cellcolor{blue!8} \textbf{\underline{0.30}} &\cellcolor{blue!8} \textbf{\underline{0.60}} &\cellcolor{blue!8} \textbf{\underline{0.63}} &\cellcolor{blue!8} \textbf{\underline{0.61}} &\cellcolor{blue!8} \textbf{\underline{0.32}} &\cellcolor{blue!8} \textbf{\underline{0.31}} &\cellcolor{blue!8} \textbf{\underline{0.31}} &\cellcolor{blue!8} \textbf{\underline{0.50}} &\cellcolor{blue!8} \textbf{\underline{0.67}} &\cellcolor{blue!8} \textbf{\underline{0.59}} \\

        \midrule[0.08em]
        \multirow{3}{*}{Map. \cite{mapanything}} 
        & VGGT-Long~\cite{vggtlong}  
        & 1.23 & 1.23 & 1.23 & 0.41 & 0.62 & 0.51 & 1.86 & 1.52 & 1.69 & 1.98 & 1.82 & 1.90 & 0.68 & 1.09 & 0.88 & \textbf{\underline{2.73}} & 1.12 & \textbf{\underline{1.92}} & \textbf{\underline{0.85}} & 0.62 & 0.73 & 1.39 & 1.15 & 1.27 \\
        & VGGT-SLAM~\cite{vggtslam}  
        & 1.66 & 9.36 & 5.51 & 2.38 & 4.74 & 3.56 & 2.28 & 9.32 & 5.80 & 4.97 & 9.54 & 7.26 & 7.69 & 7.87 & 7.78 & 9.37 & 6.80 & 8.08 & 6.20 & 7.07 & 6.63 & 4.93 & 7.81 & 6.37 \\
        
        &\cellcolor{blue!8} TALO 
        &\cellcolor{blue!8} \textbf{\underline{1.10}} &\cellcolor{blue!8} \textbf{\underline{1.14}} &\cellcolor{blue!8} \textbf{\underline{1.12}} &\cellcolor{blue!8} \textbf{\underline{0.38}} &\cellcolor{blue!8} \textbf{\underline{0.59}} &\cellcolor{blue!8} \textbf{\underline{0.49}} &\cellcolor{blue!8} \textbf{\underline{1.09}} &\cellcolor{blue!8} \textbf{\underline{0.76}} &\cellcolor{blue!8} \textbf{\underline{0.92}} &\cellcolor{blue!8} \textbf{\underline{1.65}} &\cellcolor{blue!8} \textbf{\underline{1.42}} &\cellcolor{blue!8} \textbf{\underline{1.54}} &\cellcolor{blue!8} \textbf{\underline{0.51}} &\cellcolor{blue!8} \textbf{\underline{0.92}} &\cellcolor{blue!8} \textbf{\underline{0.72}} &\cellcolor{blue!8} 3.04 &\cellcolor{blue!8} \textbf{\underline{0.93}} &\cellcolor{blue!8} 1.99 &\cellcolor{blue!8} 0.86 &\cellcolor{blue!8} \textbf{\underline{0.58}} &\cellcolor{blue!8} \textbf{\underline{0.72}} &\cellcolor{blue!8} \textbf{\underline{1.23}} &\cellcolor{blue!8} \textbf{\underline{0.90}} &\cellcolor{blue!8} \textbf{\underline{1.07}} \\
        
        \bottomrule[0.17em]
    \end{tabular}
    }
    
    \label{tab:waymo_geom}
\end{table*}

\begin{table*}[t]
    \centering
    \caption{
        Reconstructed geometry evaluation on the \textbf{nuScenes} \cite{nuScenes} dataset, reported as Accuracy, Completeness, and Chamfer Distance.
    }\vspace{-2ex}
    \tablestyle{2pt}{1.05}
    \resizebox{0.99\linewidth}!{
    \begin{tabular}{cccccccccccccccccccccccccc}
        \toprule[0.17em]
        \multirow{3}{*}{\textbf{\makecell{Foundation\\Model}}} &
        \multirow{3}{*}{\textbf{\makecell{Alignment\\Strategy}}} &
        \multicolumn{3}{c}{\textbf{scene-0003} } & 
        \multicolumn{3}{c}{\textbf{scene-0012} } & 
        \multicolumn{3}{c}{\textbf{scene-0013} } & 
        \multicolumn{3}{c}{\textbf{scene-0036} } & 
        \multicolumn{3}{c}{\textbf{scene-0039} } & 
        \multicolumn{3}{c}{\textbf{scene-0092} } & 
        \multicolumn{3}{c}{\textbf{scene-0094} } & 
        \multicolumn{3}{c}{\textbf{Avg.}} \\
        \cmidrule(lr){3-5}
        \cmidrule(lr){6-8}
        \cmidrule(lr){9-11}
        \cmidrule(lr){12-14}
        \cmidrule(lr){15-17}
        \cmidrule(lr){18-20}
        \cmidrule(lr){21-23}
        \cmidrule(lr){24-26}
        & &
        Acc.$^\downarrow$ & Com.$^\downarrow$ & Cha.$^\downarrow$ &
        Acc.$^\downarrow$ & Com.$^\downarrow$ & Cha.$^\downarrow$ &
        Acc.$^\downarrow$ & Com.$^\downarrow$ & Cha.$^\downarrow$ &
        Acc.$^\downarrow$ & Com.$^\downarrow$ & Cha.$^\downarrow$ &
        Acc.$^\downarrow$ & Com.$^\downarrow$ & Cha.$^\downarrow$ &
        Acc.$^\downarrow$ & Com.$^\downarrow$ & Cha.$^\downarrow$ &
        Acc.$^\downarrow$ & Com.$^\downarrow$ & Cha.$^\downarrow$ &
        Acc.$^\downarrow$ & Com.$^\downarrow$ & Cha.$^\downarrow$  \\
        \midrule[0.08em]

        \multirow{3}{*}{VGGT \cite{vggt}} 
        & VGGT-Long~\cite{vggtlong} 
        & \textbf{\underline{0.38}} & \textbf{\underline{1.15}} & \textbf{\underline{0.76}} & 0.58 & \textbf{\underline{1.01}} & \textbf{\underline{0.79}} & 1.02 & 1.03 & 1.02 & 1.39 & \textbf{\underline{1.66}} & \textbf{\underline{1.54}} & 1.35 & 1.70 & 1.52 & 1.12 & \textbf{\underline{0.84}} & 0.98 & \textbf{\underline{1.53}} & \textbf{\underline{0.58}} & \textbf{\underline{1.06}} & 1.05 & \textbf{\underline{1.14}} & \textbf{\underline{1.09}} \\
        & VGGT-SLAM~\cite{vggtslam}  
        & 0.97 & 1.76 & 1.36 & 8.49 & 7.39 & 7.94 & 5.48 & 6.17 & 5.82 & 8.88 & 8.98 & 8.93 & 7.52 & 7.25 & 7.38 & 4.58 & 5.51 & 5.05 & 3.46 & 1.53 & 2.50 & 5.63 & 5.51 & 5.57 \\
        &\cellcolor{blue!8} TALO 
        &\cellcolor{blue!8} 0.49 &\cellcolor{blue!8} 1.40 &\cellcolor{blue!8} 0.94 &\cellcolor{blue!8} \textbf{\underline{0.57}} &\cellcolor{blue!8} 1.10 &\cellcolor{blue!8} 0.83 &\cellcolor{blue!8} \textbf{\underline{0.85}} &\cellcolor{blue!8} \textbf{\underline{0.90}} &\cellcolor{blue!8} \textbf{\underline{0.88}} &\cellcolor{blue!8} \textbf{\underline{1.36}} &\cellcolor{blue!8} 1.72 &\cellcolor{blue!8} \textbf{\underline{1.54}} &\cellcolor{blue!8} \textbf{\underline{1.18}} &\cellcolor{blue!8} \textbf{\underline{1.59}} &\cellcolor{blue!8} \textbf{\underline{1.38}} &\cellcolor{blue!8} \textbf{\underline{1.00}} &\cellcolor{blue!8} 0.86 &\cellcolor{blue!8} \textbf{\underline{0.93}} &\cellcolor{blue!8} 1.74 &\cellcolor{blue!8} 0.80 &\cellcolor{blue!8} 1.27 &\cellcolor{blue!8} \textbf{\underline{1.03}} &\cellcolor{blue!8} 1.19 &\cellcolor{blue!8} 1.11 \\
        
        \midrule[0.08em]

        \multirow{3}{*}{$\pi^3$ \cite{pi3}} 
        & VGGT-Long~\cite{vggtlong}  
        & \textbf{\underline{0.24}} & \textbf{\underline{1.24}} & \textbf{\underline{0.74}} & \textbf{\underline{0.43}} & 1.10 & 0.76 & 1.04 & \textbf{\underline{1.40}} & 1.22 & \textbf{\underline{1.62}} & \textbf{\underline{2.43}} & \textbf{\underline{2.03}} & 1.60 & 2.17 & 1.88 & \textbf{\underline{0.60}} & \textbf{\underline{0.67}} & \textbf{\underline{0.63}} & 1.26 & 1.35 & 1.30 & 0.97 & 1.48 & 1.22 \\
        & VGGT-SLAM~\cite{vggtslam}  
        & 1.10 & 1.81 & 1.46 & 6.66 & 7.79 & 7.22 & 4.15 & 3.84 & 4.00 & 9.28 & 9.25 & 9.26 & 9.19 & 9.26 & 9.23 & 8.66 & 8.30 & 8.48 & 2.85 & 2.34 & 2.59 & 5.98 & 6.08 & 6.03 \\
        &\cellcolor{blue!8} TALO 
        &\cellcolor{blue!8} 0.27 &\cellcolor{blue!8} 1.31 &\cellcolor{blue!8} 0.79 &\cellcolor{blue!8} 0.44 &\cellcolor{blue!8} \textbf{\underline{0.99}} &\cellcolor{blue!8} \textbf{\underline{0.72}} &\cellcolor{blue!8} \textbf{\underline{0.94}} &\cellcolor{blue!8} \textbf{\underline{1.40}} &\cellcolor{blue!8} \textbf{\underline{1.17}} &\cellcolor{blue!8} 1.67 &\cellcolor{blue!8} 2.47 &\cellcolor{blue!8} 2.07 &\cellcolor{blue!8} \textbf{\underline{1.50}} &\cellcolor{blue!8} \textbf{\underline{1.98}} &\cellcolor{blue!8} \textbf{\underline{1.74}} &\cellcolor{blue!8} 0.66 &\cellcolor{blue!8} 0.75 &\cellcolor{blue!8} 0.71 &\cellcolor{blue!8} \textbf{\underline{0.90}} &\cellcolor{blue!8} \textbf{\underline{1.07}} &\cellcolor{blue!8} \textbf{\underline{0.98}} &\cellcolor{blue!8} \textbf{\underline{0.91}} &\cellcolor{blue!8} \textbf{\underline{1.42}} &\cellcolor{blue!8} \textbf{\underline{1.17}} \\

        \midrule[0.08em]
        \multirow{3}{*}{Map. \cite{mapanything}} 
        & VGGT-Long~\cite{vggtlong}  
        & 0.52 & \textbf{\underline{1.41}} & 0.97 & 0.87 & \textbf{\underline{0.99}} & 0.93 & 0.79 & 1.43 & 1.11 & 1.54 & 1.98 & 1.76 & 1.33 & 2.45 & 1.89 & 1.47 & 0.59 & 1.03 & 2.07 & 0.87 & 1.47 & 1.23 & 1.39 & 1.31 \\
        & VGGT-SLAM~\cite{vggtslam}  
        & 0.46 & 1.15 & 0.80 & 0.65 & 9.41 & 5.03 & 4.57 & 7.64 & 6.10 & 9.20 & 9.44 & 9.32 & 6.91 & 8.98 & 7.94 & 7.13 & 2.45 & 4.79 & 7.31 & 5.30 & 6.31 & 5.18 & 6.34 & 5.76 \\
        
        &\cellcolor{blue!8} TALO 
        &\cellcolor{blue!8} \textbf{\underline{0.48}} &\cellcolor{blue!8} 1.44 &\cellcolor{blue!8} \textbf{\underline{0.96}} &\cellcolor{blue!8} \textbf{\underline{0.83}} &\cellcolor{blue!8} \textbf{\underline{0.99}} &\cellcolor{blue!8} \textbf{\underline{0.91}} &\cellcolor{blue!8} \textbf{\underline{0.61}} &\cellcolor{blue!8} \textbf{\underline{1.10}} &\cellcolor{blue!8} \textbf{\underline{0.86}} &\cellcolor{blue!8} \textbf{\underline{1.35}} &\cellcolor{blue!8} \textbf{\underline{1.57}} &\cellcolor{blue!8} \textbf{\underline{1.46}} &\cellcolor{blue!8} \textbf{\underline{1.10}} &\cellcolor{blue!8} \textbf{\underline{1.65}} &\cellcolor{blue!8} \textbf{\underline{1.38}} &\cellcolor{blue!8} \textbf{\underline{1.63}} &\cellcolor{blue!8} \textbf{\underline{0.33}} &\cellcolor{blue!8} \textbf{\underline{0.98}} &\cellcolor{blue!8} \textbf{\underline{1.78}} &\cellcolor{blue!8} \textbf{\underline{0.51}} &\cellcolor{blue!8} \textbf{\underline{1.15}} &\cellcolor{blue!8} \textbf{\underline{1.11}} &\cellcolor{blue!8} \textbf{\underline{1.09}} &\cellcolor{blue!8} \textbf{\underline{1.10}} \\
        
        \bottomrule[0.17em]

    \end{tabular}
    }\vspace{-2ex}
    \label{tab:nuscenes_geom}
\end{table*}

\section{Experiments}
\label{sec:exp}
\subsection{Experimental Setup}
\noindent\textbf{Datasets.}
We conduct experiments on two multi-camera outdoor datasets: Waymo \cite{waymo} with 5 cameras and nuScenes \cite{nuScenes} with 6 cameras. 
Images are synchronized at 2 Hz and downsampled following the default settings of each foundation model, and LiDAR scans are accumulated as the same Hz as ground-truth for reconstruction evaluation.
Sky pixels are masked using a dedicated sky-segmentation model.

\noindent\textbf{Metrics. }Following common practices, we evaluate trajectories using Absolute Trajectory Error (ATE RMSE), Relative Translation Error (RTE RMSE), and Relative Rotation Error (RRE RMSE), while evaluating 3D reconstruction quality using Accuracy, Completeness, and Chamfer Distance.
Predicted camera poses are aligned to the ground truth using the Umeyama algorithm \cite{umeyama}, and the same transformation is applied to the reconstructed point clouds.

\noindent \textbf{Baselines.} We mainly compare against VGGT-SLAM~\cite{vggtslam} and VGGT-Long~\cite{vggtlong}.
Since both systems are designed exclusively for monocular video and employ only VGGT \cite{vggt} as the backbone, we re-implemented them within a unified framework using their official code, with identical back-end optimization, loop-closure mechanisms (both from VGGT-SLAM \cite{vggtslam}), and experiment settings, while extending the pipeline to support multi-camera sequences and additional foundation models including $\pi^3$ \cite{pi3} and MapAnything \cite{mapanything}.
This setup enables a clean and controlled study of alignment behavior across different datasets and backbones.

\noindent \textbf{Implementation Details.} All experiments are conducted on a single NVIDIA RTX 6000 Ada (48~GB) GPU. 
During optimization and evaluation, we retain only points whose prediction confidence is above the 60\textsuperscript{th} percentile. 
To avoid VGGT-SLAM~\cite{vggtslam}’s divergent predictions from dominating the geometry metrics, we clamp per-point errors to 10 m when evaluating 3D reconstruction quality.
The voxel resolution $\delta_v$ for control-point generation is set to 5\% of the current submap’s point-cloud radius, while the neighborhood size for control-point filtering is fixed to $Q=32$. 
The submap length is set to $L=2$ to achieve a more responsive online reconstruction (i.e., one update per second under the current 2~Hz frame rate). 
The number of overlapping frames is simply set to $O=L$. 
More implementation details are presented in the supplementary materials.



\subsection{Trajectory Evaluation}
\label{sec:exp_traj}
We evaluate camera trajectory accuracy on Waymo~\cite{waymo} (5 cameras) and nuScenes~\cite{nuScenes} (6 cameras).
As summarized in Table~\ref{tab:waymo_traj} and Table~\ref{tab:nuscenes_traj}, our proposed TALO achieves the best results across all datasets and backbones, \textbf{with zero failure cases}.  
The average ATE remains consistently around $\sim$1\,m on both datasets for all tested backbones.  
More importantly, rotational accuracy improves substantially.  
For instance, on Waymo, our method reduces RRE from 0.71$^\circ$ (VGGT-Long~\cite{vggtlong}) to only 0.14$^\circ$, achieving nearly a $\mathbf{5\times}$ improvement.  
As shown in Fig.~\ref{fig:comp}, RRE better reflects trajectory deviation than ATE.  
In the two visualized examples, VGGT-Long produces trajectories with noticeable directional drift, despite similar ATE values, while our TALO preserves the correct orientation, leading to much lower RRE.  
In contrast, the $\mathrm{SL}(4)$ alignment used in VGGT-SLAM~\cite{vggtslam} exhibits extreme instability in such outdoor, long-trajectory scenarios, failing in over 60\% of the tested sequences.  
Most failures are caused by divergence or collapse, as illustrated in Fig.~\ref{fig:comp}.  

\subsection{Geometry Evaluation}
\label{sec:exp_pcd}
\begin{figure*}[t]
  \centering
  \includegraphics[width=1\linewidth]{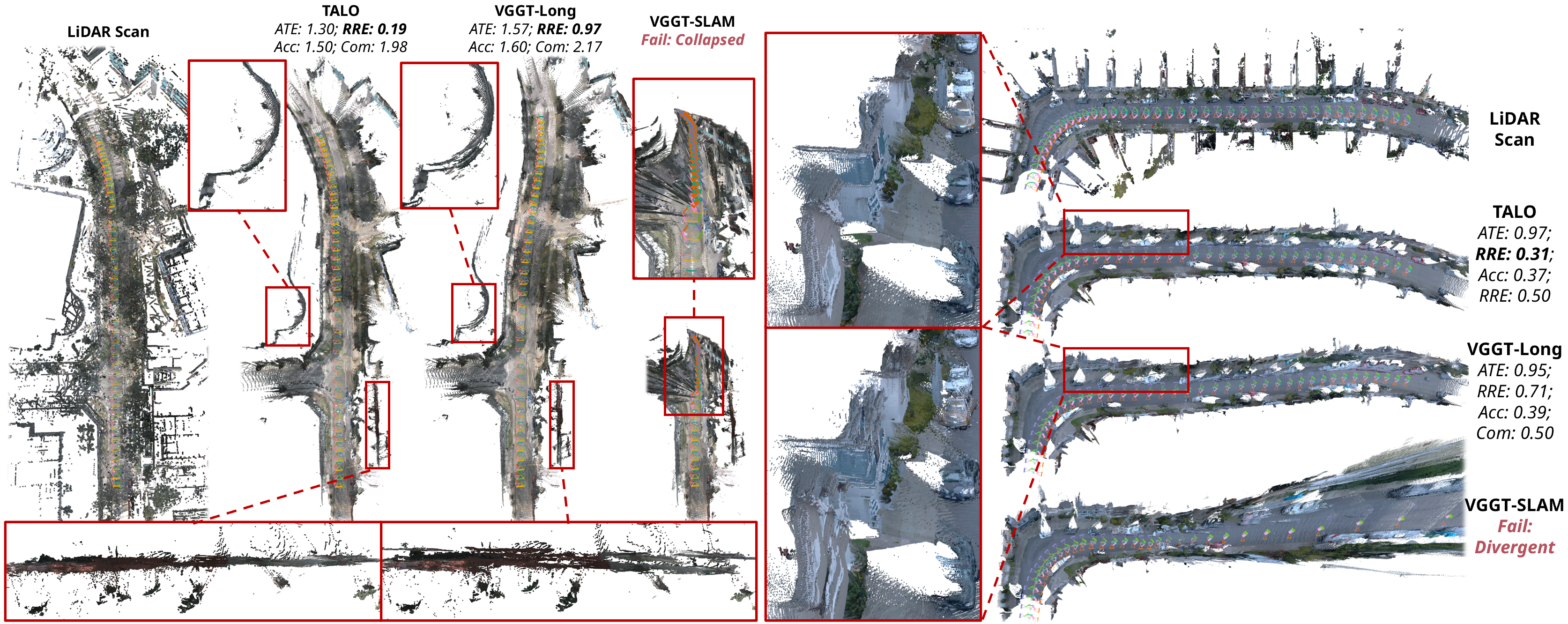}
  \vspace{-3ex}
   \caption{Qualitative comparison of the reconstructed trajectories and geometries with $\pi^3$ \cite{pi3} as backbone on \textbf{nuScenes} \cite{nuScenes} scene-0039 (\textit{left}) and \textbf{Waymo} \cite{waymo} scene 16345319168590318167 (\textit{right}), respectively. More results are shown in the supplementary materials.}
   \label{fig:comp}\vspace{-1.5ex}
\end{figure*}

We report point-cloud evaluation results on Waymo~\cite{waymo} and nuScenes~\cite{nuScenes} in Table~\ref{tab:waymo_geom} and Table~\ref{tab:nuscenes_geom}.
It can be seen that our TALO achieves the best accuracy, completeness, and Chamfer Distance across most scenes and models.
However, it is important to note that the ground-truth point clouds are obtained by accumulating LiDAR sweeps, whose capture range and coverage differ significantly from cameras.
As a result, these metrics do not reliably reflect true geometric accuracy.
For example, in the two scenes visualized in Fig.~\ref{fig:comp}, the reconstruction produced by VGGT-Long~\cite{vggtlong} exhibits clear trajectory drift, which warps the predicted geometry.
Yet because LiDAR covers a much larger spatial extent than image-based predictions, the warped point clouds still lie largely within the LiDAR coverage region, resulting in similar error scores for VGGT-Long (see Fig.~\ref{fig:comp} for exact values) despite visibly distorted geometry.

Furthermore, since bird’s-eye visualizations at hundreds-of-meters scale obscure local structure, 
we additionally include zoomed-in views in Fig.~\ref{fig:comp}.
These close-ups show that our method recovers sharper and more accurate fine-grained geometry, largely eliminating multi-layer artifacts which are frequently observed in prior methods (e.g., ghosted façades and duplicated vehicles) caused by misaligned submaps.
This improvement arises from our globally aggregated control points and the ability of TPS to correct spatially varying distortions while preserving global smoothness and local rigidity.
Although these misaligned surfaces constitute only a \textit{minor} fraction of the entire scene and are \textit{weakly} reflected in metrics, they are \textit{crucial} for safety-critical tasks such as distance perception.

\begin{figure}[t]
  \centering
  \includegraphics[width=\linewidth]{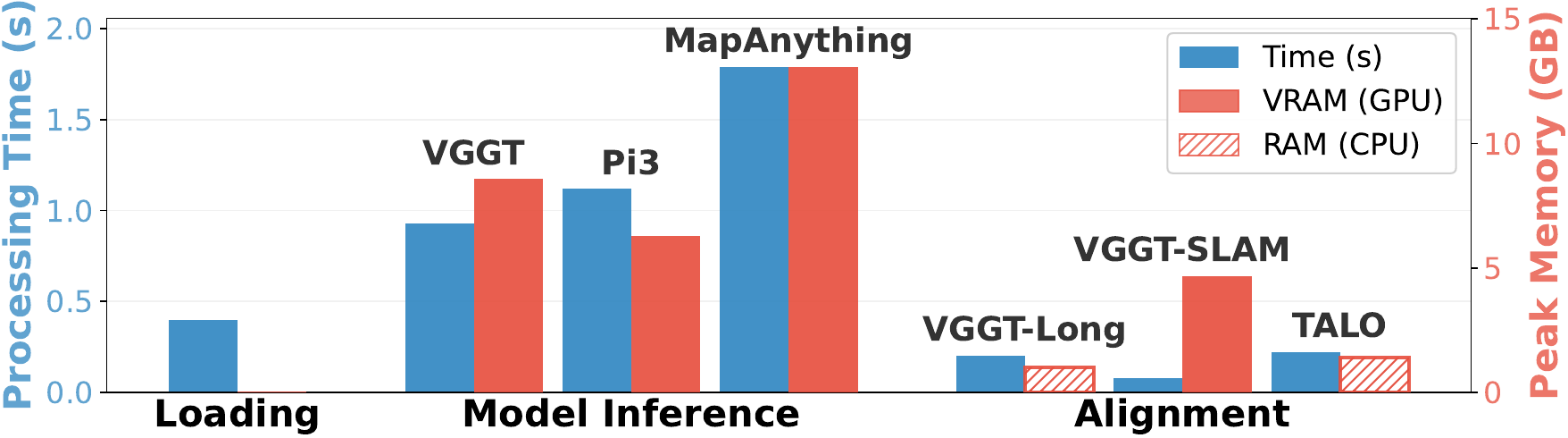}
   \label{fig:onecol}
   \vspace{-3ex}
   \caption{Per-submap processing time and peak memory usage.}
\vspace{-3ex}
\end{figure}

\subsection{Efficiency Study}
We report the per-submap processing time and peak memory usage on Waymo. 
As shown, TALO introduces only a minor overhead in time and memory compared to VGGT-Long, which is negligible relative to model inference that dominates the runtime cost. 
In practical deployments where loading, inference, and alignment are executed in parallel, none of the alignment methods constitute a system bottleneck. 
TALO and VGGT-Long are implemented in NumPy and therefore report RAM usage, whereas VGGT-SLAM performs GPU-based parallel searches over hundreds of SL(4) hypotheses, leading to substantially higher VRAM consumption. 
This analysis confirms that TALO achieves a favorable trade-off between accuracy and efficiency.

\subsection{Ablation Study}

To further understand the behavior of our system, we conduct ablations on Waymo~\cite{waymo} based on VGGT~\cite{vggt}.
In addition to the surround-view setting, we evaluate a monocular configuration using only the front camera.
We compare TALO with a classic SLAM pipeline, DROID-SLAM~\cite{droid}, as well as two online DUSt3R-like reconstruction methods, CUT3R~\cite{cut3r} and MASt3R-SLAM~\cite{mast3rslam}.
Trajectory and geometry results are reported in Table~\ref{tab:waymo_mono_traj} and Table~\ref{tab:waymo_mono_geom}, respectively.
It can be seen that TALO achieves the best trajectory accuracy and reconstruction quality across nearly all scenes.

\begin{table}[t]
    \centering
    \caption{\textbf{Monocular} \textit{camera} trajectory accuracy on \textbf{Waymo} \cite{waymo}, reported as ATE (RMSE [m]). Best results are \textbf{\underline{bold underlined}}.}
    \vspace{-1ex}
    \tablestyle{2pt}{1.05}
    \resizebox{0.99\linewidth}!{
    \begin{tabular}{lccccccccc|c}
        \toprule[0.17em]
        \textbf{Method} & 16345. & 18382. & 31561. & 34618. & 40584. & 52001. & 61045. & Avg.  \\
        \midrule[0.08em]
        DROID-SLAM \cite{droid}  & 3.71 & 0.30 & 0.45 & 8.65 & 7.62 & 4.17 & 0.26 & 3.59\\
        MASt3R-SLAM \cite{mast3rslam}  & 4.50 & 0.56 & 1.83 & 12.54 & 1.41 & 5.43 & 1.20 & 3.92\\
        CUT3R \cite{cut3r} & 8.78 & 3.81 & 5.79 & 24.02 & 7.26 & 13.21 & 3.23   & 9.44 \\
        VGGT-Long \cite{vggtlong}  & 1.65 & 0.22 & 0.37 & \textbf{\underline{3.62}} & 1.08 & 1.85 & \textbf{\underline{0.10}} & 1.27\\
        \rowcolor{blue!8}TALO  & \textbf{\underline{0.71}} & \textbf{\underline{0.19}} & \textbf{\underline{0.22}} & 3.74 & \textbf{\underline{0.87}} & \textbf{\underline{1.63}}  & 0.20 &\textbf{\underline{1.08}} \\

        \bottomrule[0.17em]

    \end{tabular}
    }
    \label{tab:waymo_mono_traj}
\end{table}

\begin{table}[t]
\vspace{-1ex}
    \centering
    \caption{\textbf{Monocular} \textit{geometry} evaluation on \textbf{Waymo} \cite{waymo}, reported as Chamfer Distance. Best results are \textbf{\underline{bold underlined}}.}
    \vspace{-1ex}
    \tablestyle{2pt}{1.05}
    \resizebox{0.99\linewidth}!{
    \begin{tabular}{lccccccccc|c}
        \toprule[0.17em]
        \textbf{Method} & 16345. & 18382. & 31561. & 34618. & 40584. & 52001. & 61045. & Avg.  \\
        \midrule[0.08em]
        DROID-SLAM \cite{droid}  & 2.70 & 5.69 & 3.89 & 5.53 & 5.84 & TL & 7.44 & 5.18\\
        MASt3R-SLAM \cite{mast3rslam}  & 2.45 & 3.14 & 2.92 & 3.84 & 2.03 & 5.42 & 4.60 & 3.49\\
        CUT3R \cite{cut3r} & 5.92 & 5.25 & 5.58 & 6.13 & 3.87 & 7.31 & 5.37   & 5.63 \\
        VGGT-Long \cite{vggtlong}  & 1.57 & 2.23 & 2.46 & 2.10 & 1.85 & \textbf{\underline{1.80}} & 1.69 & \textbf{\underline{1.96}}\\
        \rowcolor{blue!8}TALO  & \textbf{\underline{1.53}} & \textbf{\underline{2.17}} & \textbf{\underline{2.43}} & \textbf{\underline{2.03}} & \textbf{\underline{1.80}} & 2.14  & \textbf{\underline{1.63}} &\textbf{\underline{1.96}} \\

        \bottomrule[0.17em]

    \end{tabular}
    }
    \vspace{-2ex}
    \label{tab:waymo_mono_geom}
\end{table}

\begin{figure}[t]
    \centering

    \includegraphics[width=0.48\linewidth]{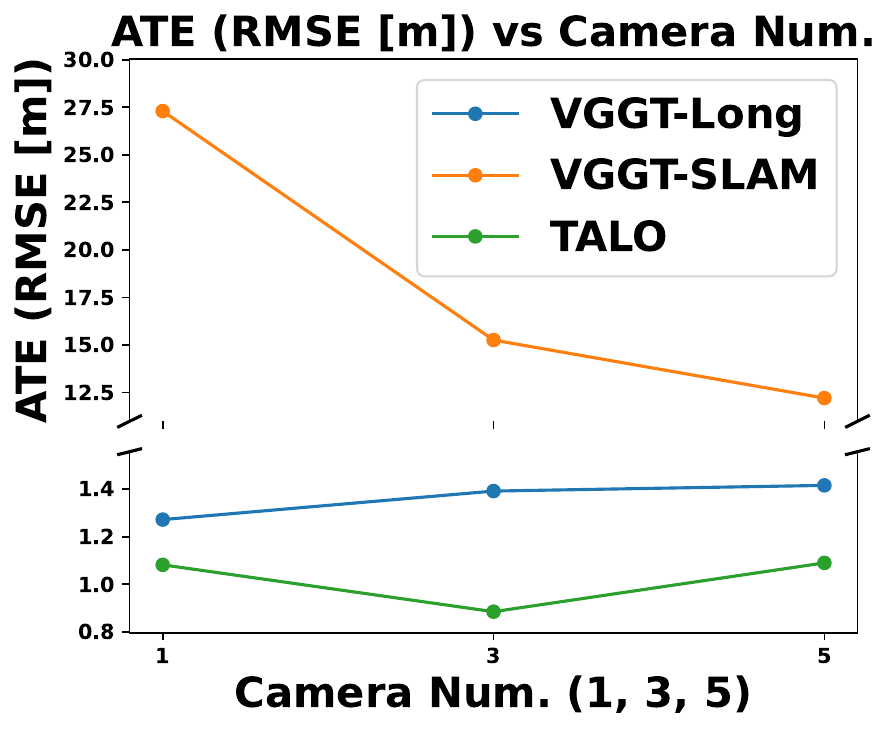}
    \hfill
    \includegraphics[width=0.48\linewidth]{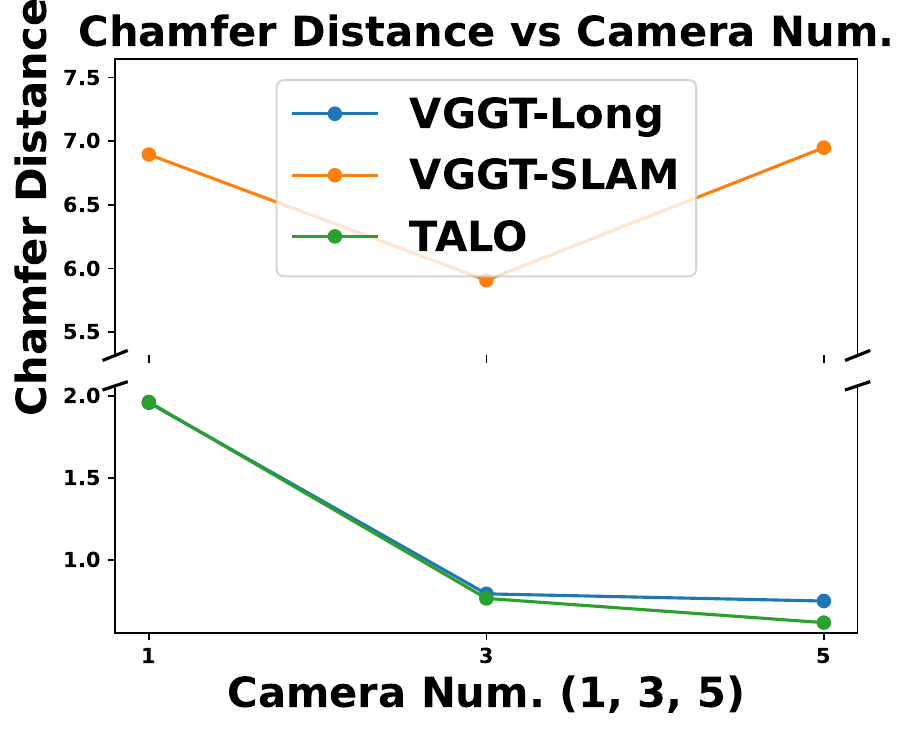}
    \includegraphics[width=0.48\linewidth]{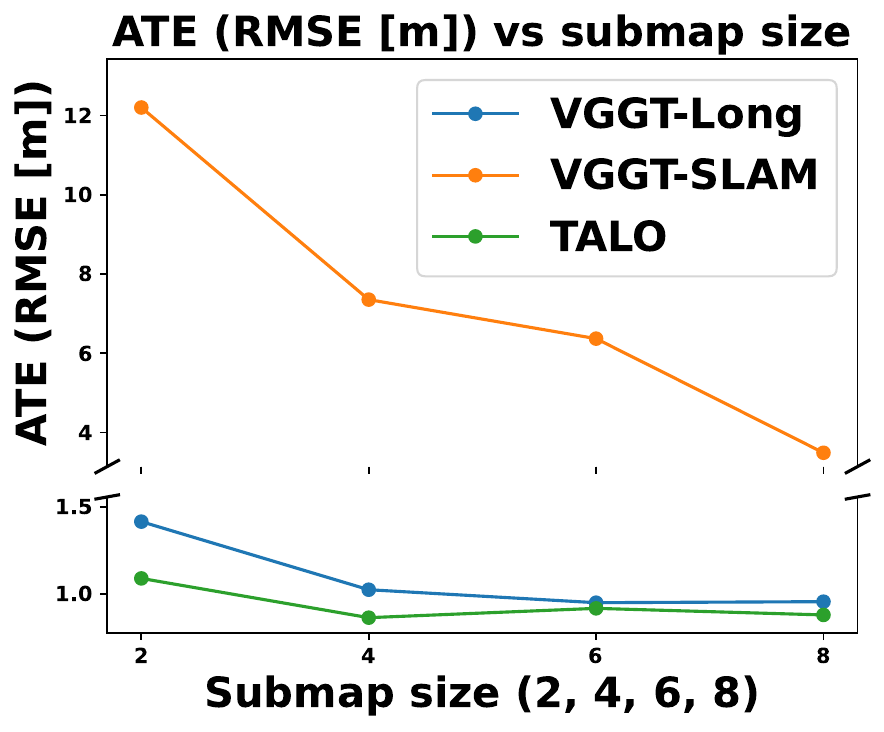}
    \hfill
    \includegraphics[width=0.48\linewidth]{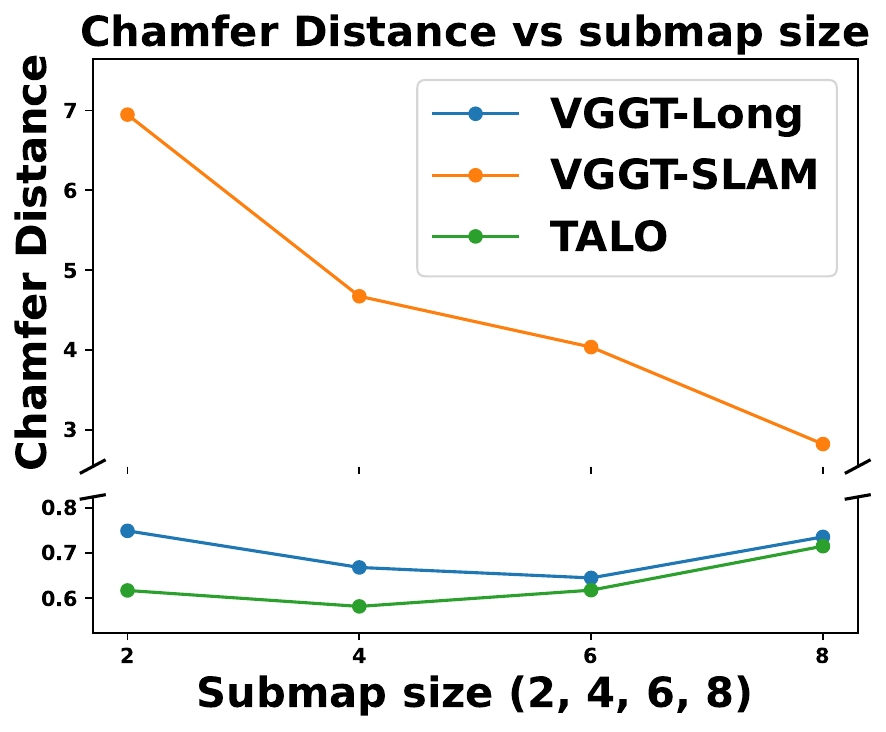}
    \caption{Effect of camera count and submap size on trajectory and geometry accuracy.
Across all settings, TALO consistently achieves the lowest ATE and Chamfer Distance. 
}\vspace{-3ex}
    \label{fig:abcd}
\end{figure}

We further analyze the effects of camera count and submap size; results are summarized in Fig.~\ref{fig:abcd}.
Both VGGT-Long~\cite{vggtlong} ($\mathrm{Sim}(3)$) and our method remain relatively stable under varying camera numbers and submap sizes, although several trends can be observed:
(1) larger submaps lead to lower trajectory error, and
(2) additional cameras improve geometric accuracy.
This likely arises because foundation models benefit from increased viewpoint diversity, producing more stable and accurate geometry predictions.
Importantly, our method consistently outperforms VGGT-Long under all tested configurations.
For VGGT-SLAM~\cite{vggtslam} ($\mathrm{SL}(4)$), we observe that both trajectory and geometry errors generally decrease as the number of cameras and the submap size increase.
We attribute this to the richer viewpoints reducing noise in the foundation model’s geometry predictions, which stabilizes the $\mathrm{SL}(4)$ estimation and lowers the risk of catastrophic divergence.


\vspace{-1ex}
\section{Conclusion}

We revisit online reconstruction with 3D vision foundation models and identify the limitations of existing global alignment strategies. 
To address this, we propose TALO, a long-term alignment framework that globally propagates sparse control points and applies Thin Plate Spline deformation for spatially adaptive corrections. 
Combined with a point-agnostic submap registration strategy, TALO serves as a plug-and-play module compatible with diverse foundation models and camera setups. 
Extensive experiments demonstrate consistent improvements in geometric consistency and trajectory accuracy across datasets.

\section{Acknowledgment}
This work was partially supported by ARC DE240100105, DP240101814, DP230101196, BA24006, and ARC Industrial Transformation Research Hubs IH230100013.


{
    \small
    \bibliographystyle{ieeenat_fullname}
    \bibliography{main}
}

\clearpage
\setcounter{page}{1}
\maketitlesupplementary

\begin{figure*}[t]
  \centering
  \includegraphics[width=1\linewidth]{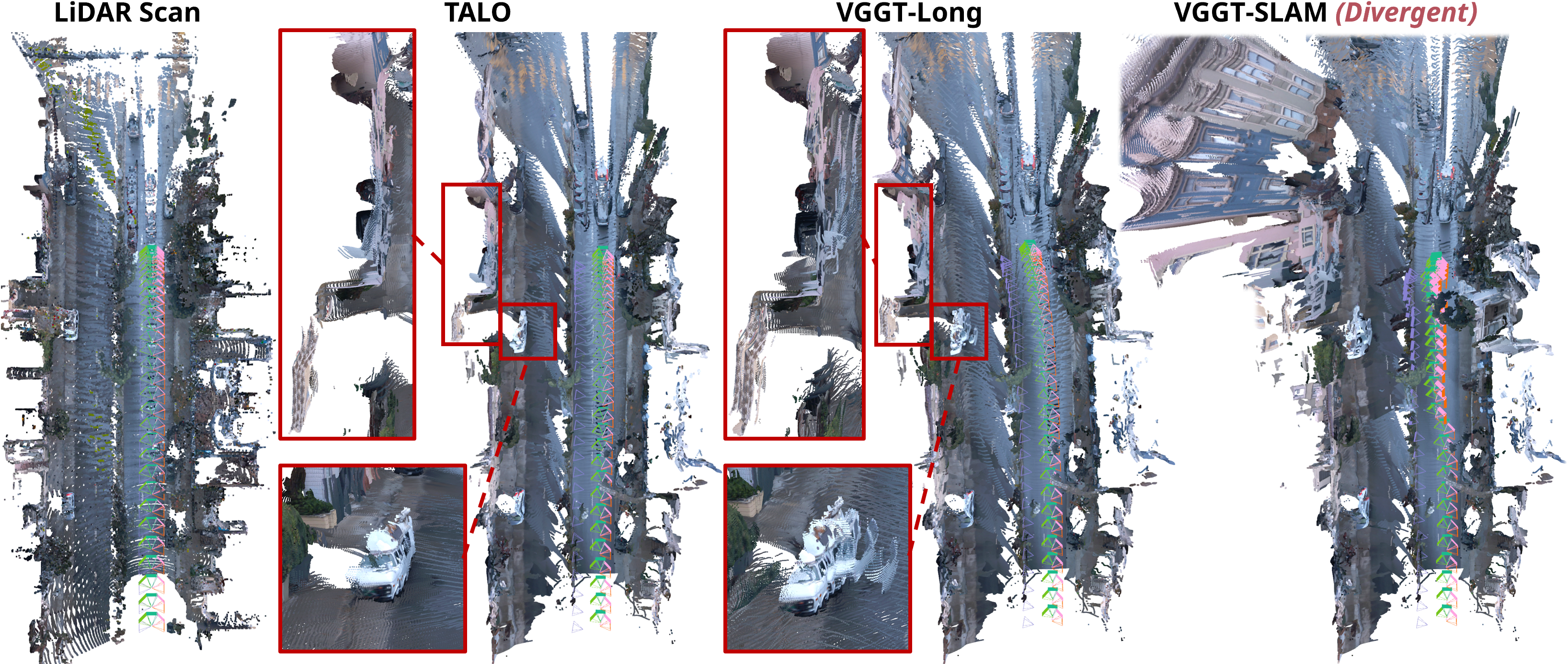}
   \caption{Qualitative comparison with VGGT \cite{vggt} on \textbf{Waymo} \cite{waymo} scene 6104545334635651714.}
   \label{fig:vggt_way}
\end{figure*}

\begin{figure*}[t]
  \centering
  \includegraphics[width=1\linewidth]{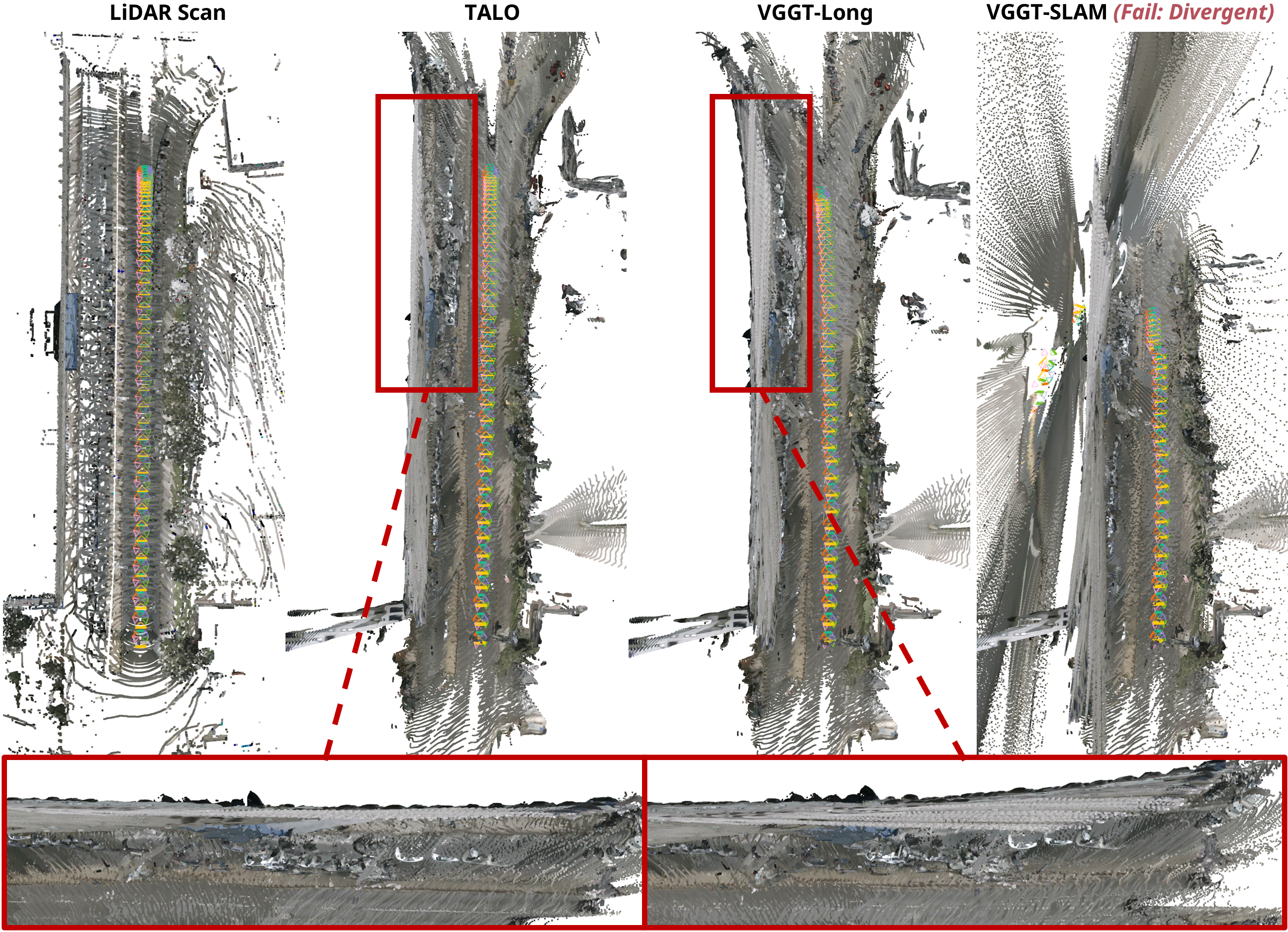}
   \caption{Qualitative comparison with VGGT \cite{vggt} on \textbf{nuScenes} \cite{nuScenes} scene-0094.}
   \label{fig:vggt_nu}
\end{figure*}

\begin{figure*}[t]
  \centering
  \includegraphics[width=0.93\linewidth]{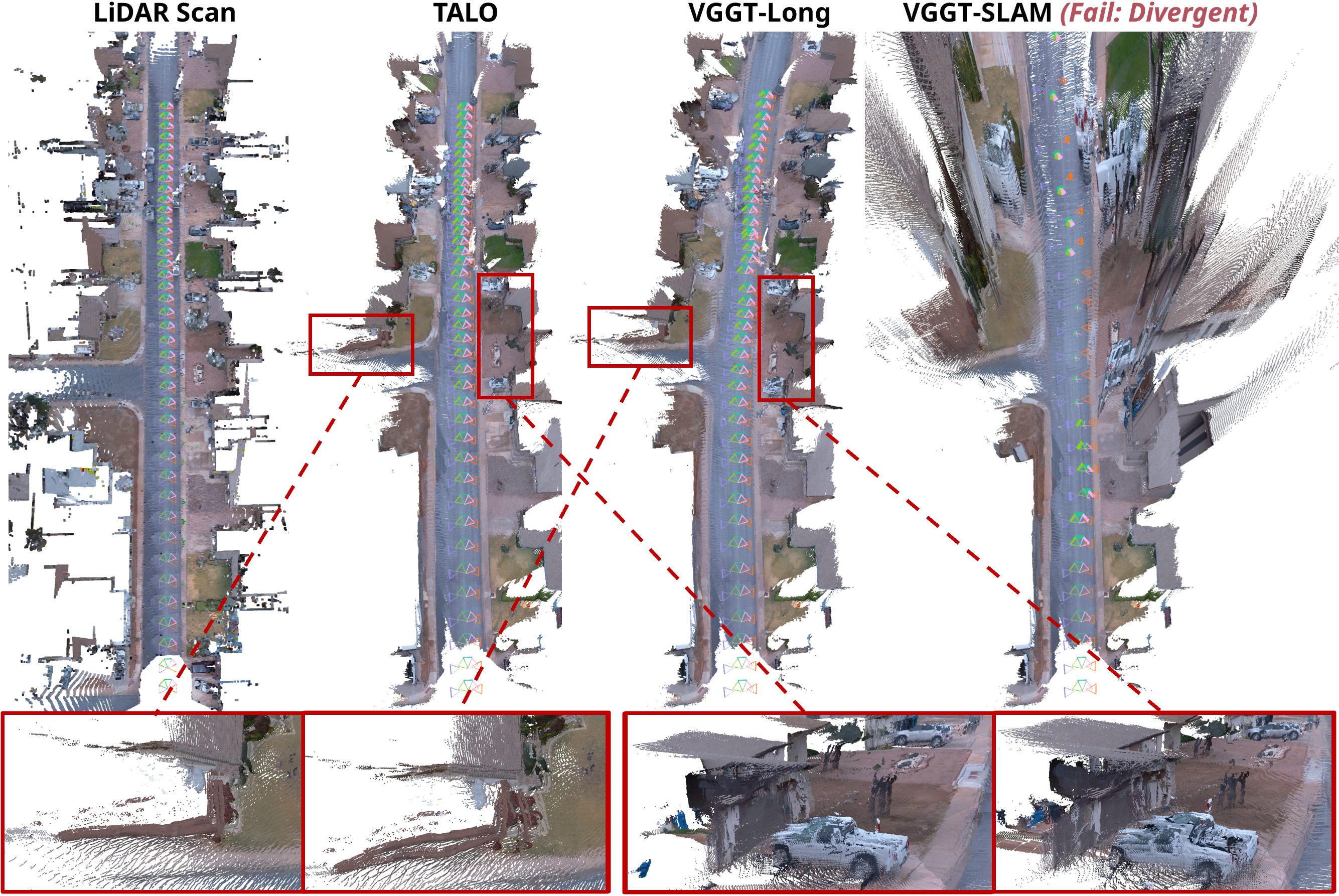}
   \caption{Qualitative comparison with $\pi^3$ \cite{pi3} on \textbf{Waymo} \cite{waymo} scene 3156155872654629090.}
   \label{fig:pi_way}
\end{figure*}

\begin{figure*}[t]
  \centering
  \includegraphics[width=0.93\linewidth]{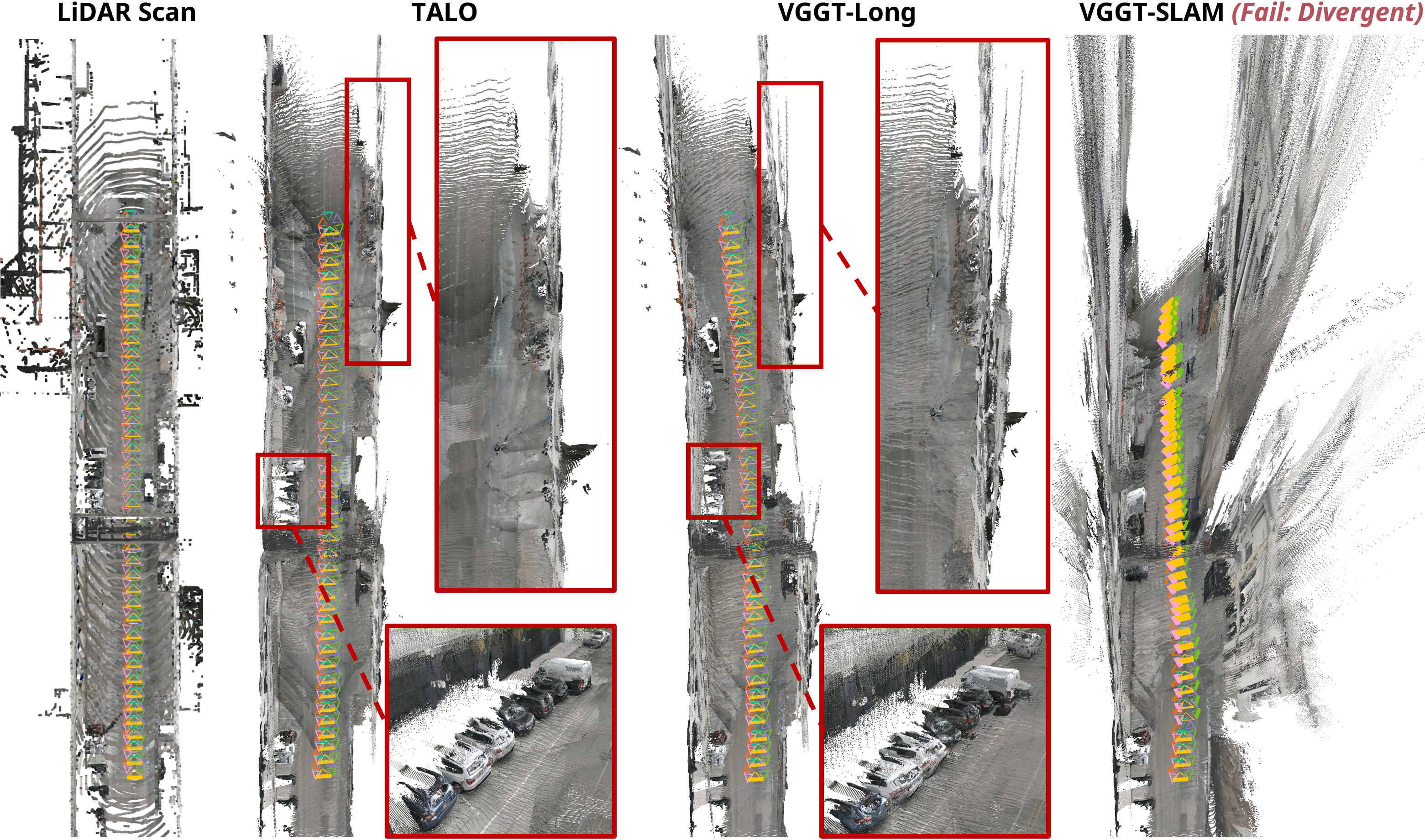}
   \caption{Qualitative comparison with $\pi^3$ \cite{pi3} on \textbf{nuScenes} \cite{nuScenes} scene-0092.}
   \label{fig:pi_nu}
\end{figure*}

\appendix

\section{Linear Transformation Assumptions}
This section analyzes the hidden assumptions behind adopting a linear transformation between two submaps: different predictions differ only by a global scale in depth ($\mathrm{Sim}(3)$) and/or by intrinsics that can be corrected through a projective warp ($\mathrm{SL}(4)$). 
Specifically, consider two camera reconstructions $\{\tilde{\mathbf{u}}_1, \mathbf{K}_1, \mathbf{D}_1, \mathbf{T}_1\}$ and $\{\tilde{\mathbf{u}}_2, \mathbf{K}_2, \mathbf{D}_2, \mathbf{T}_2\}$ observing the same scene, where $\tilde{\mathbf{u}}=(u,v,1)$ denotes homogeneous pixel coordinates, $\mathbf{K}$ the intrinsics, $\mathbf{D}$ the predicted depth, and $\mathbf{T}=[\mathbf{R}\mid \mathbf{t}]$ the extrinsics.
The 3D point $\mathbf{P}$ can be obtained by back-projection:
\begin{equation}
\mathbf{P} \;=\; [\mathbf{R}\mid \mathbf{t}]^{-1}\big(\mathbf{D}(u,v)\,\mathbf{K}^{-1}\tilde{\mathbf{u}}\big).
\end{equation}
Now, if we wish to solve for the transformation $\mathbf{H}$ relating $\mathbf{T}_1$ and $\mathbf{T}_2$ through $\mathbf{P}_1$ and $\mathbf{P}_2$, three typical cases arise:
\begin{itemize}
    \item Case 1: Consistent intrinsics and scaled depths.
    If $\mathbf{K}_1 = \mathbf{K}_2$ and $\mathbf{D}_2(u,v) = s\,\mathbf{D}_1(u,v)$ for a global scalar $s$, then the two reconstructions differ only by a uniform scale.
    This simplifying assumption is implicitly adopted by VGGT-Long~\cite{vggtlong}.
    There exists a similarity transformation $\mathbf{T} = [\,s\,\mathbf{R}\mid \mathbf{t}\,]\in\mathrm{Sim}(3)$ such that $\mathbf{P}_2 = s\,\mathbf{R}\,\mathbf{P}_1 + \mathbf{t}$, 
    which perfectly aligns the two reconstructions. 

    \item Case 2: Inconsistent intrinsics and scaled depths.
    When $\mathbf{K}_1 \neq \mathbf{K}_2$ and $\mathbf{D}_2(u,v) = s\,\mathbf{D}_1(u,v)$, the change in intrinsics induces a global projective distortion that a single Sim(3) cannot capture.
    This assumption is adopted by VGGT-SLAM~\cite{vggtslam}.
    Nevertheless, under the projective reconstruction theorem, there always exists a homogeneous linear mapping 
    $\mathbf{H} \in \mathrm{SL}(4)$ that transforms one reconstruction into the other such that
    $\mathbf{P}_1 = \mathbf{H}\,\mathbf{P}_2$.

    \item Case 3: Nonlinear depth distortion.
    When the depth predictions deviate nonlinearly across spatial locations, 
    i.e., $\mathbf{D}_2(u,v) = f(\mathbf{D}_1(u,v))$, where $f(\cdot)$ is non-linear and spatially varying, 
    no global linear transformation can exactly align the two reconstructions, regardless of whether the internal calibration is consistent.
    Any linear alignment forces the optimizer to overfit one region at the expense of another, and bend the trajectory to compensate for local depth inconsistencies.
\end{itemize}

This analysis exposes the hidden assumptions underlying the two point-based submap alignment methods and explains their fundamental limitations. 
The more severely the corresponding assumption is violated, the poorer the alignment performance becomes.
In scenarios with large submap size and few inter-submap alignments, the violation is often mild and aligns more closely with Case~1 and Case~2, allowing VGGT-Long \cite{vggtlong} and VGGT-SLAM \cite{vggtslam} to perform reasonably well.
However, realistic outdoor multi-camera settings generally correspond to Case~3, where the assumptions in Case 1 and Case 2 rarely hold.
As a result, enforcing a global rigid or projective transformation inevitably leaves residual inconsistencies and distorts the trajectory.

\section{Additional Implementation Details}

For models such as VGGT~\cite{vggt} and $\pi^3$~\cite{pi3}$\!$ which lack metric-scale, we estimate a global scale factor before control-point propagation to avoid using high-DOF TPS for global scale correction. 
For models like MapAnything~\cite{mapanything}$\!$ which maintain metric scale prediction, this step is omitted, and we also disable scale estimation in VGGT-Long’s~\cite{vggtlong} $\mathrm{Sim}(3)$ alignment for fairness, whereas in VGGT-SLAM~\cite{vggtslam} ($\mathrm{SL}(4)$ formulation) the scale cannot be explicitly decoupled.

TALO makes \textbf{no assumptions} about the number of cameras, and therefore naturally supports \textbf{arbitrary} setups ranging from monocular and stereo to surround-view systems.
To exploit the fixed inter-camera baselines in multi-camera configurations, we introduce a simple rig averaging strategy.
Specifically, for each non-reference camera, we collect its predicted relative poses to the reference camera across time, and estimate a single time-invariant rig transform by averaging rotation and translation separately: rotations are averaged on $\mathrm{SO}(3)$ using a chordal $\ell_2$ mean, implemented by summing rotation matrices and projecting the result back to $\mathrm{SO}(3)$ via SVD, while translations are averaged in Euclidean space.
By aggregating these relative poses over time, rig averaging (\textbf{RA}) suppresses frame-wise prediction noise and enforces consistency with the underlying rigid camera setup.
This simple yet effective strategy is applied fairly to both VGGT-SLAM and VGGT-Long.

%

\section{Additional Qualitative Comparisons}
We provide additional qualitative comparisons in Fig.~\ref{fig:vggt_way}, \ref{fig:vggt_nu}, \ref{fig:pi_way}, and \ref{fig:pi_nu}.
As shown, VGGT-Long \cite{vggtlong} ($\mathrm{Sim}(3)$) exhibits noticeable trajectory drift in long-range sequences, together with local misalignments that manifest as multi-layer artifacts in the reconstructed geometry.
VGGT-SLAM \cite{vggtslam} ($\mathrm{SL}(4)$) is often disrupted by noise in the predicted point clouds, leading to severe geometric divergence or complete reconstruction failure.
In contrast, TALO produces trajectories that closely match the ground truth and yields locally well-aligned geometry with clean, artifact-free details.

\end{document}